\documentclass[runningheads]{llncs}
\usepackage[T1]{fontenc}
\usepackage{lmodern} 
\usepackage{hyperref}
\usepackage{color}

\usepackage{amsmath}
\usepackage{float}

\usepackage{graphicx}
\usepackage{wrapfig}
\usepackage{cite}
\usepackage[utf8]{inputenc}
\usepackage{epsfig}
\usepackage{multirow}
\usepackage{makecell}
\usepackage{adjustbox}

\title{GuidedSAM-Plume: An expert-guided segmentation framework for volcanic plumes}

\titlerunning{GuidedSAM-Plume}

\author{
Taddeo D'Adamo\inst{1} \and
Simon Thivet\inst{2} \and
Catherine Pothier\inst{1} \and
Riccardo Simionato\inst{2,3} \and
Allan Fries\inst{2} \and
Jonathan Lemus\inst{2,3} \and
Costanza Bonadonna\inst{2}\and
Laure Tougne Rodet\inst{1} \and
Bertrand Kerautret\inst{1}
}

\authorrunning{T. D'ADAMO et al.}

\institute{
Université Lumière Lyon 2, INSA Lyon, CNRS, Universite Claude Bernard Lyon 1, Ecole Centrale de Lyon, LIRIS, UMR5205, 69029 Bron, France \and
Department of Earth Sciences, University of Geneva, Geneva, Switzerland \and
Department of Computer Sciences, University of Geneva, Carouge, Switzerland
}

\begin{document}

\maketitle
\setcounter{footnote}{0}

\begin{abstract}

Quantitative analysis of environmental phenomena increasingly relies on image-based measurements. In volcanology, the geometric characterization of volcanic plumes and clouds from RGB images remains hindered by the scarcity of dedicated segmentation solutions and partially handcrafted annotations remain widely prevalent. We present \emph{GuidedSAM-Plume}, an open-source expert-guided framework tailored for the interactive segmentation of volcanic plumes and clouds. It leverages the zero-shot, spatial prompt-based capabilities of SAM2. The framework was thoroughly validated on a challenging dataset through dedicated segmentation metrics and field-relevant volcanological descriptors. From minimal user inputs, it is able to significantly increase performance compared to fully automatic dedicated solutions while preserving the reliability of derived volcanological parameters. The absence of retraining requirements lowers adoption barriers within the volcanological community, and the framework's demonstrated suitability for complex, diffuse plume-shaped phenomena extends its applicability to broader environmental science domains. The re-annotated dataset, produced under the guidance of expert volcanologists within the proposed framework, ultimately yields a more standardized release. The framework and accompanying dataset are made available at
 \url{https://github.com/tad-damo/GuidedSAM-Plume}.
 

\end{abstract}

\section{Introduction}

Across scientific domains, image segmentation is a core task enabling downstream analysis of natural phenomena \cite{Szeliski2022}. In fields from biomedical imaging \cite{Ronneberger2015} to environmental monitoring \cite{Wilkes2022}, binary masks are a critical prerequisite for quantitative characterization of observed objects. This work aims at leveraging recent computer vision advances to benefit environmental sciences, with a concrete application to volcanology. Explosive volcanic eruptions produce ash plumes and clouds, influencing atmospheric processes and impacting environments from local to global scales.
Segmentation is of crucial importance in that field as delineating plumes allows for the quantitative analysis and retrieval of real-world physical parameters, such as plume morphology, spatial extent, and temporal dynamics \cite{Simionato2022a}. However, automated plume segmentation in visible imagery remains limited due to scarce contributions \cite{Wilkes2022, giffard2025vigia} and deep learning solutions have not yet reached widespread adoption in this domain. Traditional threshold-based algorithms are sensitive to contrast conditions, leading many volcanologists to rely on partly handcrafted masks. In these conditions, obtaining segmentation masks not only requires considerable annotation effort but also leads to inconsistencies across datasets produced by different operators, due to a lack of methodological standardization. This introduces subjective biases when comparing studies \cite{Benenson_2019_CVPR}. 
Meanwhile, research in computer vision has demonstrated strong capabilities for the segmentation task  \cite{Sultana2020} and has progressed toward foundation models \cite{Bommasani2022}, achieving high generalization performance across a wide range of applications, as exemplified by the Segment Anything Model (SAM) \cite{Kirillov2023}. We were led by the hypothesis that, despite the inherent complexity of volcanic plumes, such advances should render their analysis more tractable even in data-limited settings.

\medskip

\noindent The main contributions of this work are as follows:

\begin{enumerate}
	
	\item \textbf{Efficient interactive segmentation framework for volcanic plumes and clouds}  
We introduce \emph{GuidedSAM-Plume}, an open-source interactive segmentation framework designed through quantitative analysis and close collaboration with volcanologists to address actual operational needs.  It enables reliable and reproducible plume delineation in both images and video sequences, maintaining refinement control while substantially reducing annotation costs, by leveraging the zero-shot capabilities of SAM2 \cite{Ravi2024} within an expert-in-the-loop strategy. The eliminated requirement for extensive retraining should lower adoption barriers for volcanology and broader environmental monitoring applications facing similar challenges.

	\item \textbf{Standardized dataset} A novel dataset is released comprising 47 observational sequences totaling 759 annotated frames drawn from multiple eruptive events, representative of data typically employed in volcanological studies. 
	Initially to evaluate the framework, the dataset was re-annotated under the supervision of expert volcanologists following the structured workflow. The resulting segmentation masks demonstrated adequate quality for downstream analysis and being inherently more standardized than those produced by multiple independent annotators, both annotation versions have been made publicly available.

	\item \textbf{Validation for real-world science}  
	The effectiveness of the framework is demonstrated through segmentation quality assessment against the manually annotated ground truth of the aforementioned dataset. Beyond standard metrics, outputs are further validated via application-driven evaluation based on the extraction of geometric and physical plume descriptors, supporting both the validity of the proposed framework and the relevance of the standardized dataset for real-world volcanological applications.
	
\end{enumerate}

\noindent The segmentation framework \emph{GuidedSAM-Plume} as well as the dataset have been made publicly available at 
 \url{https://github.com/tad-damo/GuidedSAM-Plume}.

\section{Related Work}

\subsection{Volcanology-dedicated solutions}
While visible-wavelength cameras are widely deployed on volcanoes for
monitoring purposes \cite{Aravena2025, Simionato2022a}, dedicated
image segmentation solutions for the scientific analysis of volcanic
eruptions remain limited. Early semi-automatic approaches, such as the
Plume Ascent Tracker \cite{valade2014plume} and PlumeTraP
\cite{Simionato2022a} require substantial manual effort and are
strongly dependent on operator input for segmentation. Initial deep
learning approaches remained volcano-specific, targeting individual
sites such as Mount Etna (Italy)\cite{guerrero2022convolutional} or Sabancaya (Peru)
\cite{centeno2024near}. Wilkes et al. \cite{Wilkes2022} attempted to
address this limitation with three UNet- \cite{Ronneberger2015} or
FPN-based \cite{lin2017feature} architectures designed to generalize
across multiple volcanoes without retraining; however, the limited
size of their training and validation datasets suggests that reported
performance may overestimate real-world applicability. More recently,
VigiaPlume-Net \cite{giffard2025vigia} leverages DINOv2
\cite{oquab2023dinov2} for fine-grained six-category
segmentation. However, the highly variable imaging conditions and dynamic irregular plume structures limit fully automated approaches for static image segmentation and should motivate methods exploiting video sequences for better spatiotemporal characterization.
Furthermore, while other scientific domains benefit from dedicated segmentation platforms, for instance AnnotatorJ \cite{hollandi2020annotatorj} within the widely adopted ImageJ ecosystem \cite{schneider2012nih} for microscopy, no comparable solution currently exists for volcanic plume analysis.
\subsection{Interactive Segmentation}
During the last decade, deep learning has become the dominant paradigm for image segmentation, progressing from convolutional \cite{Long2015, Chen2017} to transformer-based architectures \cite{Dosovitskiy2021, cheng2021mask2former} with remarkable performance gains across wide range of applications.  Fully-automated approaches require large annotated datasets, costly training, and offer limited output control. Interactive segmentation has emerged as a compelling alternative \cite{Huang_2023_ICCV, sofiiuk2022reviving, wong2024scribbleprompt}, offering a better balance between annotation effort and output quality through iterative refinement, as demonstrated by frameworks such as Ilastik \cite{berg2019ilastik}. Interactive segmentation offers a favorable balance between annotation effort and output quality through iterative refinement, making it particularly attractive for complex, expert-guided tasks \cite{Benenson_2019_CVPR}. The more recent emergence of foundation models further transformed the field by introducing large-scale models designed to generalize across tasks and domains \cite{Dosovitskiy2021, Radford2021, Bommasani2022}. In this context, the Segment Anything Model (SAM) represents a major step toward universal image segmentation.  It enables zero-shot segmentation without task-specific training, relying on flexible user prompts such as bounding boxes, points, or coarse masks \cite{Kirillov2023}. Its architecture combines a vision transformer-based image encoder, a prompt encoder, and an efficient mask decoder. SAM has been widely adopted and has inspired numerous derivatives \cite{pachitariu2025cellpose, xie2024moving, sam_hq, zou2023segment}. Its successor, SAM2, further improves spatial coherence and additionally enables segmentation videos thank the addition of a memory bank and memory attention module \cite{Ravi2024}. More recently, SAM3 \cite{carion2026sam} goes beyond simple spatial prompting by introducing textual referring segmentation and exemplar prompts.

\section{GuidedSAM-Plume}


\begin{figure}[tbp]
	\centering
	\includegraphics[width=\textwidth]{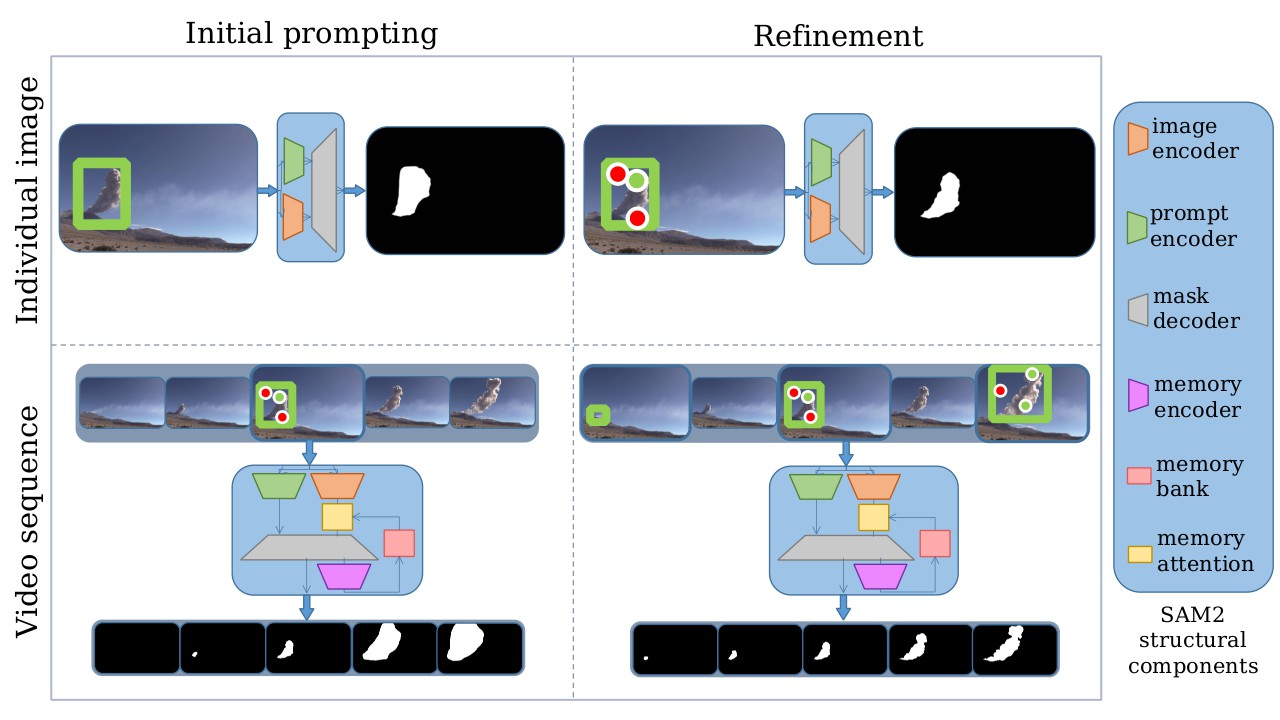}
	\caption{Overview of the \emph{GuidedSAM-Plume} framework and its workflow. For still images, segmentation is initialized with a bounding box (top left) and iteratively refined using foreground and background point prompts (top right). For video sequences, the process is initialized on a representative keyframe near the middle of the sequence (bottom left), followed by temporal tracking and refinement through additional keyframes (bottom right).}
	\label{fig:workflow}
\end{figure}

\noindent The \emph{GuidedSAM-Plume} framework follows a standardized
interactive procedure designed to minimize user effort while
preserving expert control. The expert-guided workflow of the framework
is presented in Fig. \ref{fig:workflow}. The system supports both
individual images and image sequences: in the case of single images,
each frame requires user interaction, whereas for sequences, tracking
functionalities significantly reduce the level of manual input. The
guidance signal is provided in the form of spatial prompts, including
bounding boxes as well as positive points within the plume and negative points
in the surrounding background. These prompts can be combined to enable
flexible refinement. Refinement constitutes a central component of the
framework. Owing to the inherent complexity of volcanic plume imagery,
segmentation outputs may not fully capture the physically meaningful
plume extent. In such cases, the operator can iteratively refine the
segmentation by providing additional sparse annotations. The
segmentation mask is updated accordingly until it accurately
represents the observed plume geometry. For temporal sequences,
refinement can be further improved through interaction with additional
keyframes, thereby enhancing tracking performance. As presented later
in Section~\ref{prompting_result}, quantitative analyses indicate that
an effective strategy consists in initializing the interaction with a
bounding box enclosing the target plume, followed by incremental
refinement through the addition of point-based prompts to correct
observed errors. Regarding image time series or video data, it is
recommended to initialize the workflow on a representative frame
located near the middle of the sequence due to the high variability of plume structures
over time.

\medskip
\noindent{\textbf{Workflow interface}}
\noindent To support an efficient workflow in real-world applications, the \emph{GuidedSAM-Plume} framework required to be released along with a dedicated operational interface. 
\begin{figure}[htbp]
	\centering
	\includegraphics[width=\textwidth]{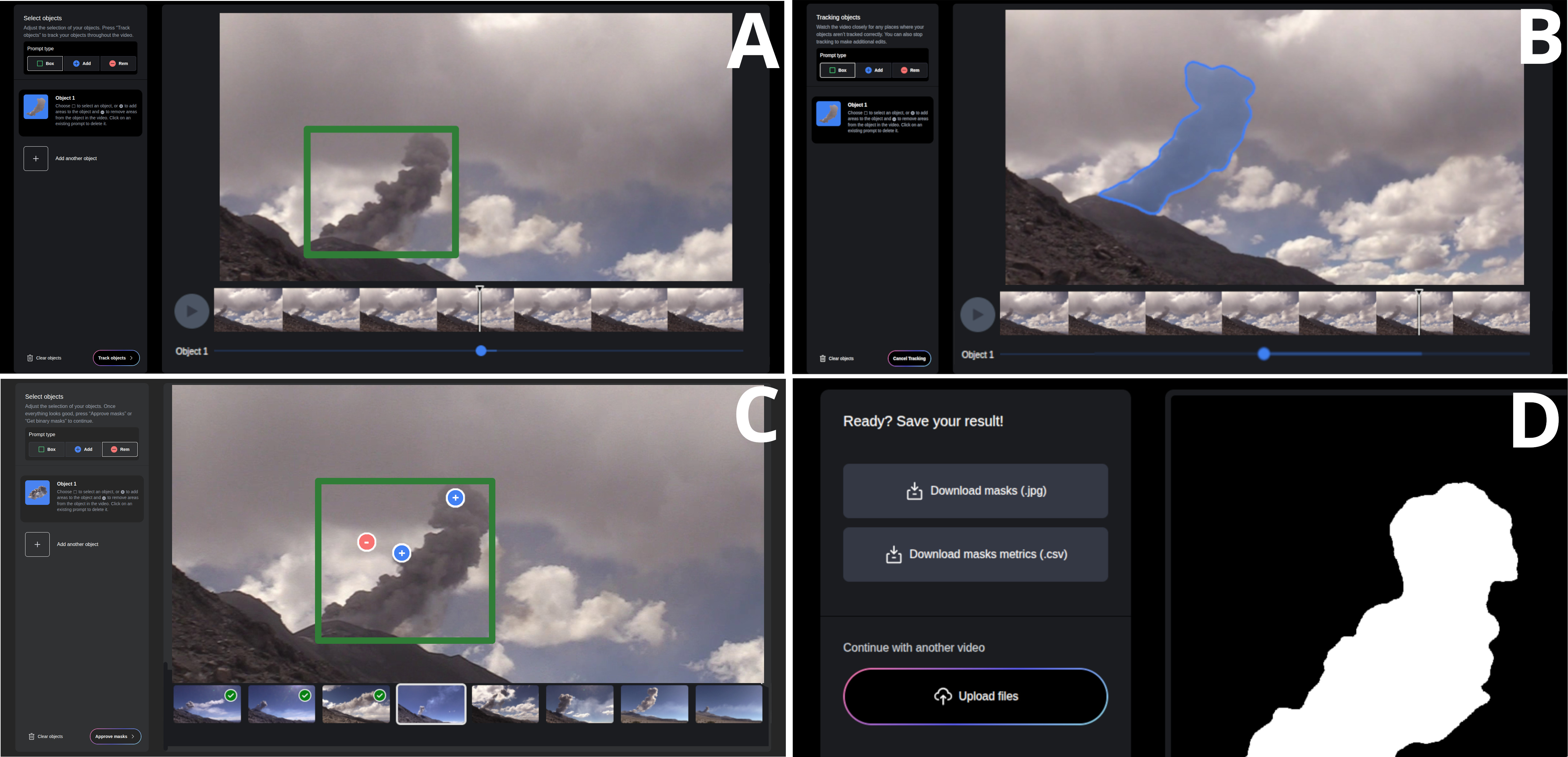}
	\caption{Presentation of the \emph{GuidedSAM-Plume} interface and its workflow. It includes keyframe prompting for sequences (A), mask propagation across frames (B), handling of individual images (C), and data export functionality (D).}
	\label{fig:guidedsamplume}
\end{figure}
Built upon the open-source \emph{SAM2 Demo} codebase \cite{Ravi2024}, it advances beyond existing SAM2 interfacing platforms \cite{SAM2_GUI, segment_anything_2_ui, sam2_studio}, offering a dedicated, fully functional interface specifically tailored for volcanic plume and cloud analysis. Its design is intentioned to support real-world usability. Notable added features include still image and sequence segmentation, multi-type prompt integration, segmentation export along with volcanology-specific metrics computation. An overview of the interface is shown in Figure~\ref{fig:guidedsamplume}, and full technical details are provided in the open-source repository. This workflow interface completes the framework that effectively integrates expert knowledge into the interpretation of complex and ambiguous plume structures.

\section{Evaluation Method}

\subsection{Protocol}\label{protocol}

\subsubsection{SAM2 assessment}
Effort was devoted to first validate a SAM2-based framework, and ensuring that it is reliable for the analysis of volcanic plume imagery. SAM2 inferences were performed using the largest variant version 2.1; the mask with highest confidence score was systematically selected among the three candidate outputs \cite{Ravi2024}. The proposed framework was benchmarked against available literature solution, namely the three architectures from \cite{Wilkes2022} and VigiaPlume-Net \cite{giffard2025vigia}, with the latter evaluated on its \emph{eruption column} class. Final masks produced through expert volcanologist guidance within the proposed framework were additionally evaluated.

Further, realistic maximal expectations were estimated determining the minimal error owed to the intrinsic architecture. Segmentation models commonly produce output logits at a reduced resolution relative to the input; in SAM2, this downsampling factor is fourfold. They are subsequently upscaled to the original image resolution using bilinear interpolation and then thresholded to generate binary masks\cite{Ravi2024}. To quantify the associated loss in performance, the ground truths were successively downscaled to the input resolution (\(1024 \times 1024\)), then to logits dimensions (\(256 \times 256\)) before being upscaled to the original image dimensions, following the same resolution transformations as in the internal processing of SAM2.

\subsubsection{Workflow establishment}
Once the reliability of zero-shot capabilities from SAM2 were validated, different workflows were assessed to determine conditions leading to optimized prompting. Experiments were conducted for static images by observing the response of the model to different combination of bounding box and point-based input prompts. The rough mask prompting capability of SAM2 \cite{Ravi2024} was not considered, as it requires extensive annotation effort and is less suited for lightweight workflows. A maximum of five prompts per image were evaluated. The effects of randomly placed points and spatially predetermined points, such as the brightness point of the distance map, were compared. In random settings, metrics were averaged over 5 independent draws. Beyond static images, SAM2 supports the propagation of user-defined prompts across frames, enabling temporally consistent segmentation of video sequences from minimal initial input \cite{Ravi2024}. This capability was evaluated by supplying manually annotated masks as prompts on selected keyframes and assessing propagation quality on the remaining frames, thereby enabling both the evaluation of video propagation performance and the identification of an optimal keyframe prompting strategy. Additional analyses were performed to characterize the impact of using the framework on volcanological applications. These aspects are discussed further in Section~\ref{volcanoimpact}.

\subsection{Metrics}

First, standard computer vision metrics derived from the counts of true positives (TP), true negatives (TN), false positives (FP) and false negatives (FN) were employed. Specifically, we report accuracy (A), precision (P), recall (R), dice coefficient (D), jaccard index (J), f1-score computed at one-pixel contours (F) and the $J\&F$ measure, average of the $J$ region similarity and $F$ contour measure \cite{Perazzi2016}. They are defined as:

\begin{gather*}
	\mathrm{A} = \frac{\mathit{TP} + \mathit{TN}}{\mathit{TP} + \mathit{TN} + \mathit{FP} + \mathit{FN}}, \qquad
	\mathrm{P} = \frac{\mathit{TP}}{\mathit{TP} + \mathit{FP}}, \qquad
	\mathrm{R} = \frac{\mathit{TP}}{\mathit{TP} + \mathit{FN}}, \\
	\mathrm{D} = \frac{2 \cdot P \cdot R}{P + R}, \qquad
	\mathrm{J} = \frac{\mathit{TP}}{\mathit{TP} + \mathit{FP} + \mathit{FN}}, \qquad
	\mathrm{F} = D_{\mathrm{Contour}}, \qquad
	\mathrm{J\&F} = \frac{J + F}{2}.
\end{gather*}

\noindent Second, to consider the relevance of segmentation results for volcanological applications during the assessment, geometric and morphological descriptors are extracted from the masks to quantify deviations from handcrafted annotations in downstream analyses. We consider aspect ratio (AR), convexity (CVX), and solidity (SLD) \cite{Simionato2024a}, defined as:

\begin{gather*}
	\mathrm{AR} = \frac{\mathit{FD}_{\min}}{\mathit{FD}_{\max}}, \qquad
	\mathrm{CVX} = \frac{\mathit{CP}_\mathrm{CH}}{\mathit{CP}_\mathrm{M}}, \qquad
	\mathrm{SLD} = \frac{\mathit{SA}_\mathrm{M}}{\mathit{SA}_\mathrm{CH}}.
\end{gather*}

\noindent where $\mathit{FD}_{\max}$ and $\mathit{FD}_{\min}$ are the maximum and minimum Feret diameters \cite{liu2015optimising}, while $\mathit{CP}_\mathrm{CH}$ and $\mathit{SA}_\mathrm{CH}$ denote contour perimeter and surface area of the convex hull, and $\mathit{CP}_\mathrm{M}$ and $\mathit{SA}_\mathrm{M}$ those of the mask. 

From these parameters, we introduce new metrics defined as the complement of the absolute relative error. The complement is taken for consistency with other evaluation metrics, such that higher values uniformly indicate better performance. Specifically, for a given parameter $X$, the metric is defined as:
\begin{gather*}
X\epsilon = 1 - \frac{|X_{\text{prediction}} - X_{\text{ground truth}}|}{X_{\text{ground truth}}}.
\end{gather*}

\noindent We report $\mathit{AR}{\epsilon}$, $\mathit{CVX}{\epsilon}$, and $\mathit{SLD}{\epsilon}$. These metrics are used to provide an agreement for volcanology parameters between predictions and ground truth.
\subsection{Dataset}

\begin{figure}[tbp]
	\centering
	\begin{adjustbox}{width=\columnwidth}
		 \begin{tabular}{ccc@{\hspace{0.02\textwidth}}ccc}
			\includegraphics[width=0.15\textwidth]{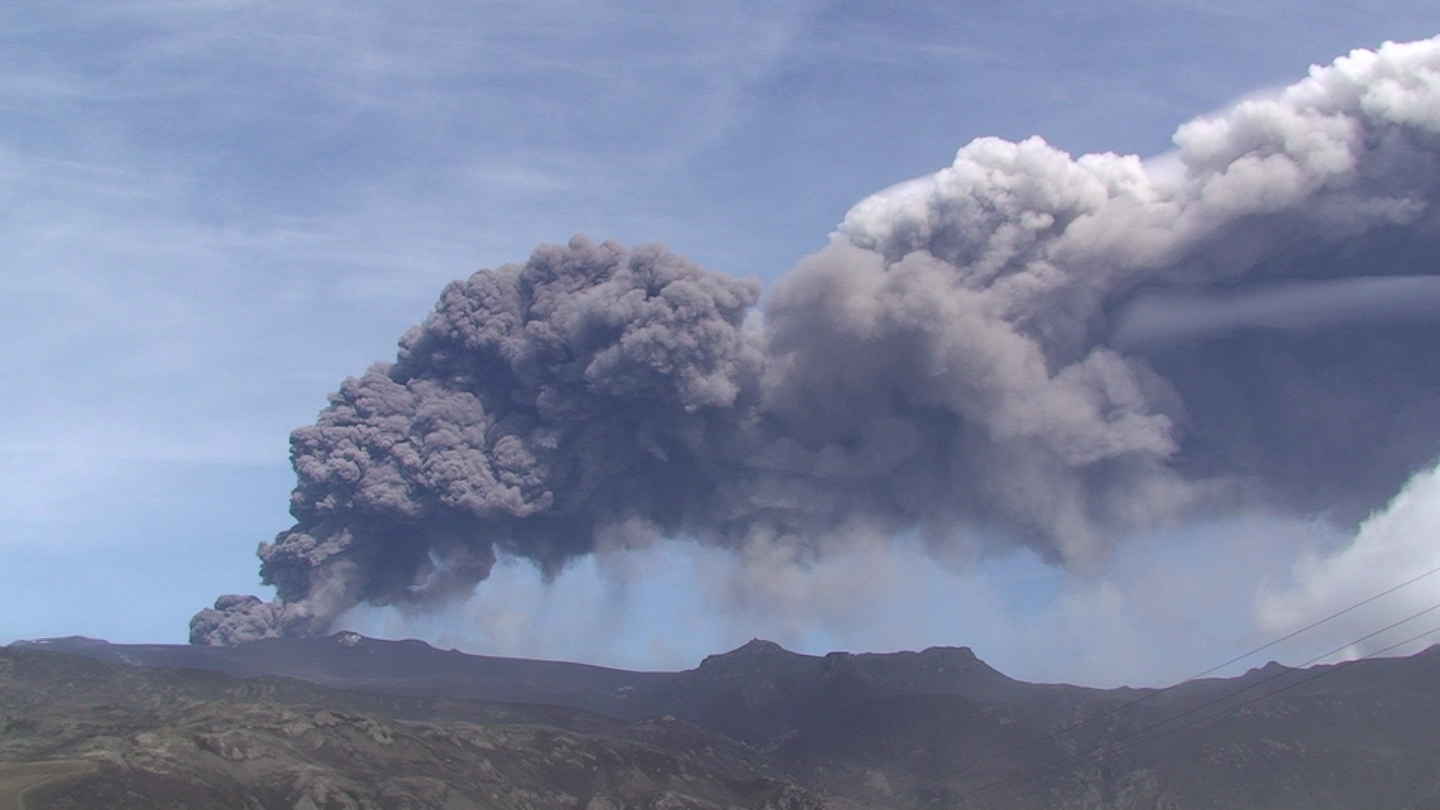}&
			\includegraphics[width=0.15\textwidth]{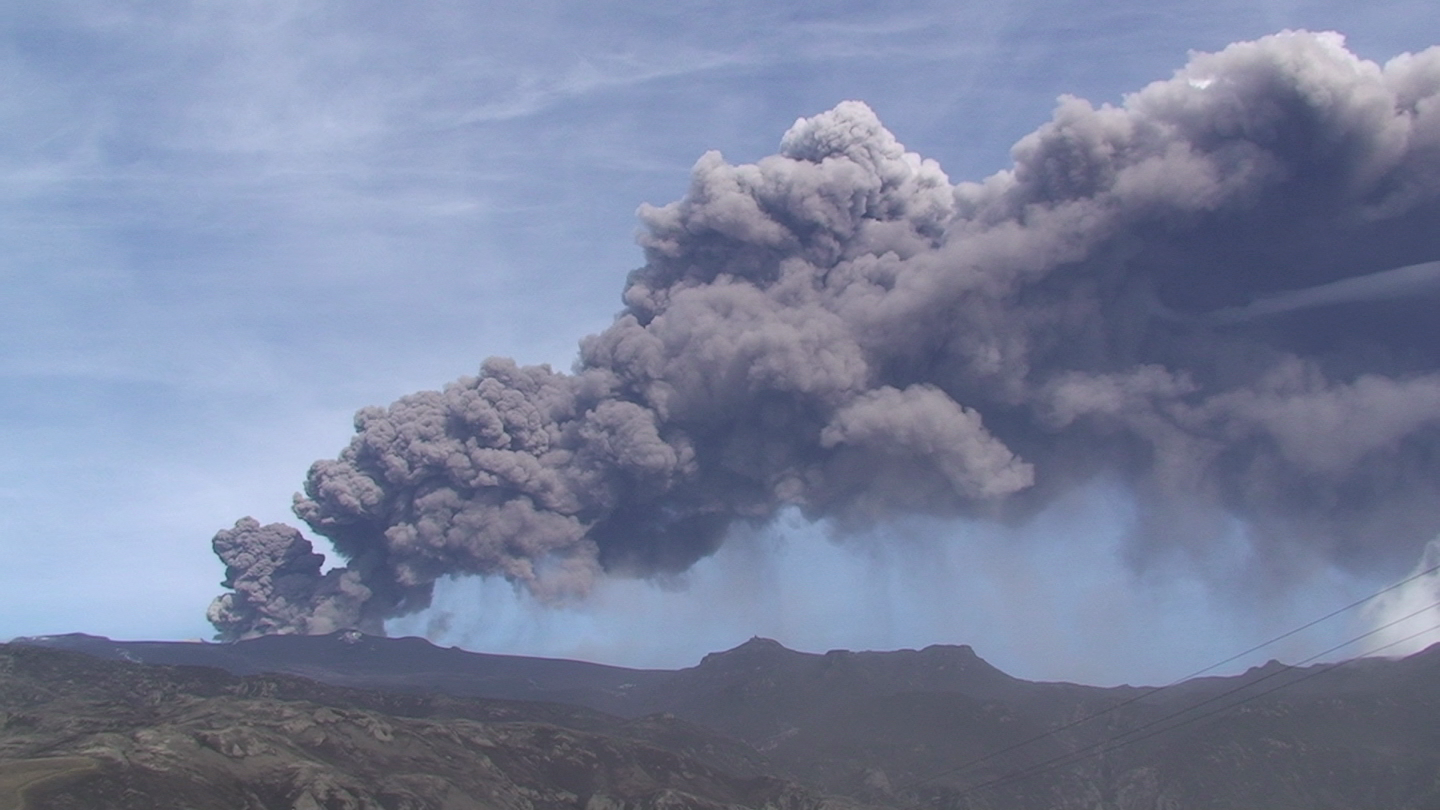}&
			\includegraphics[width=0.15\textwidth]{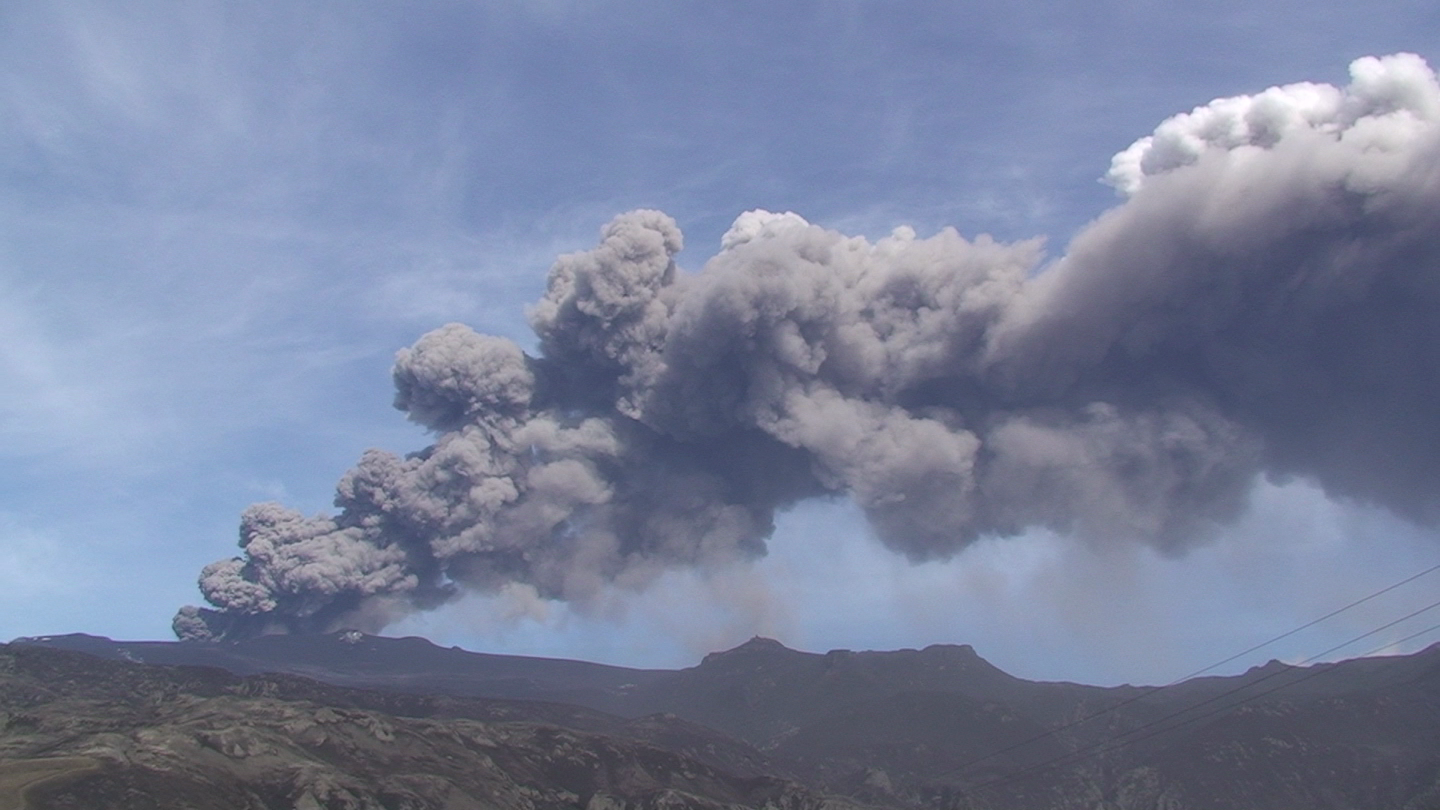}&
			\includegraphics[width=0.15\textwidth]{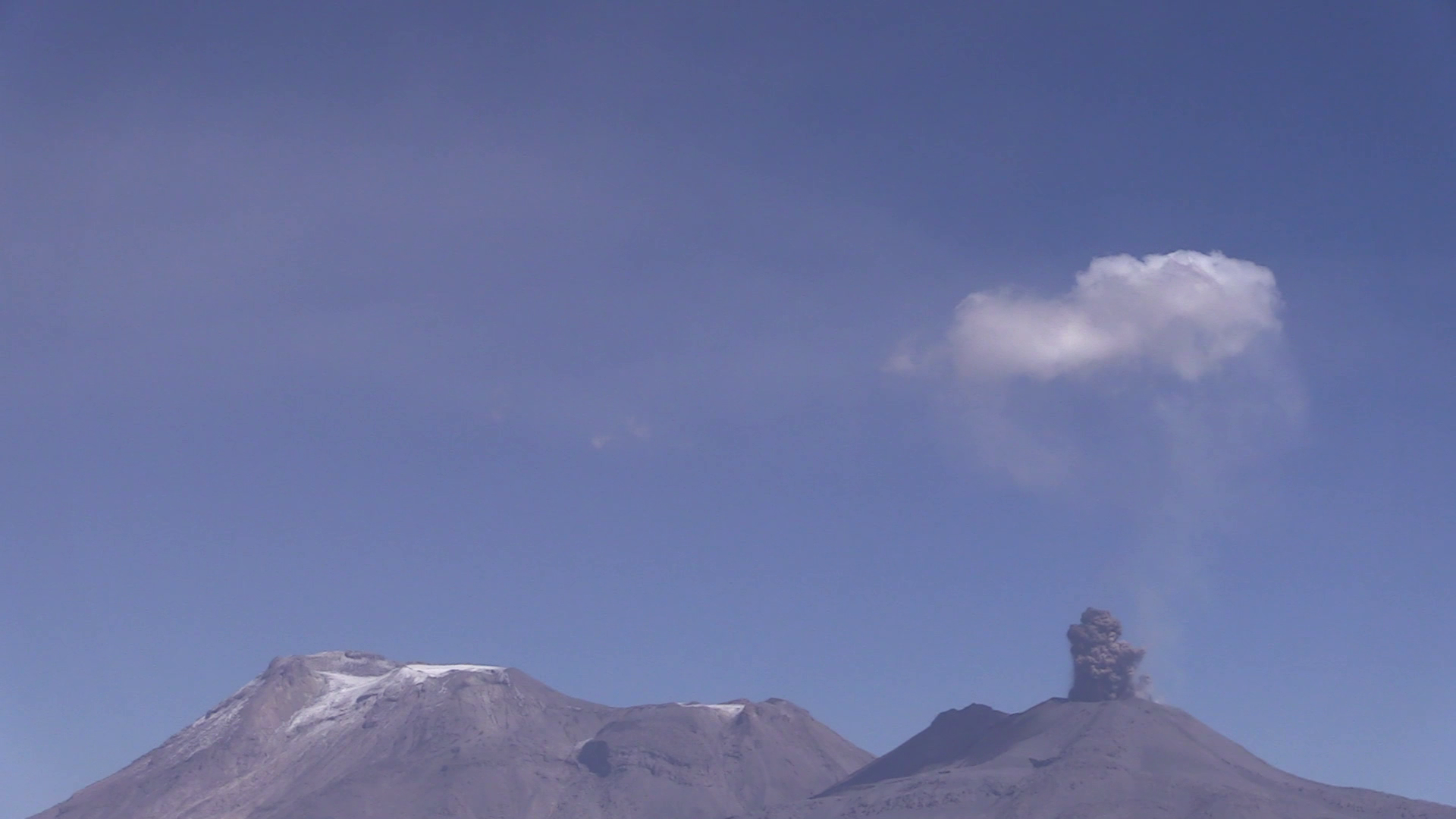}&
			\includegraphics[width=0.15\textwidth]{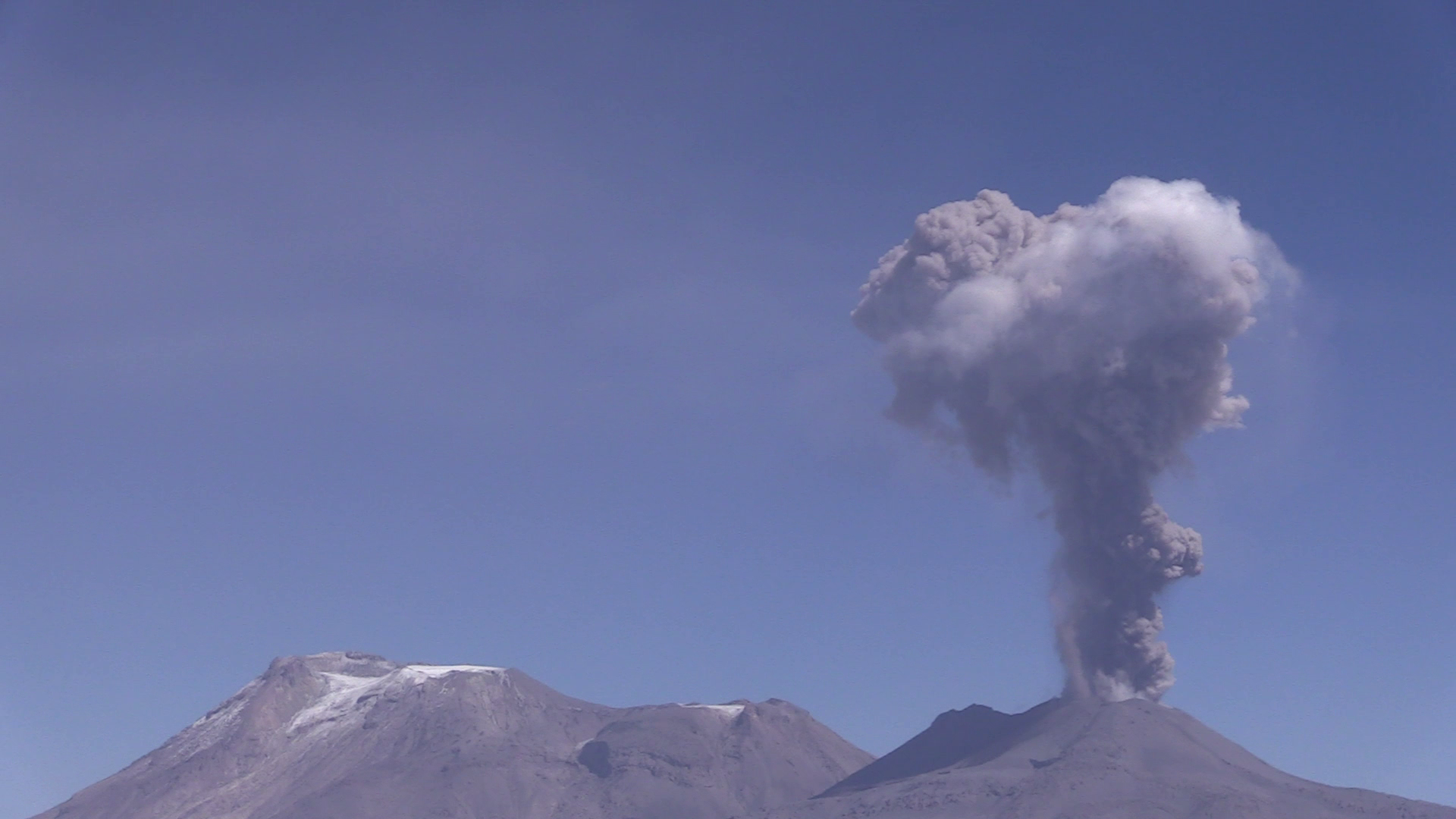}&
			\includegraphics[width=0.15\textwidth]{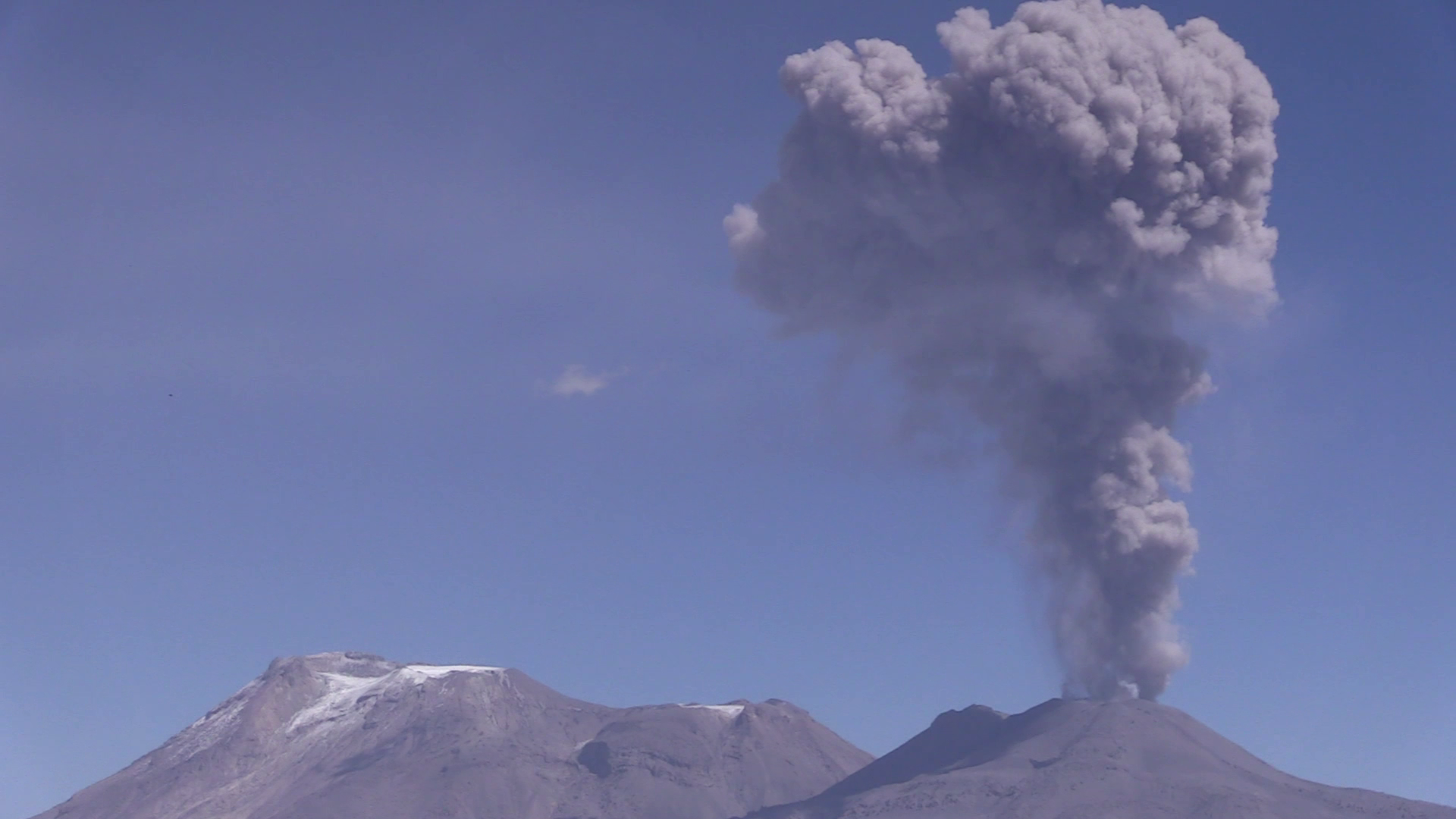}\\
			\includegraphics[width=0.15\textwidth]{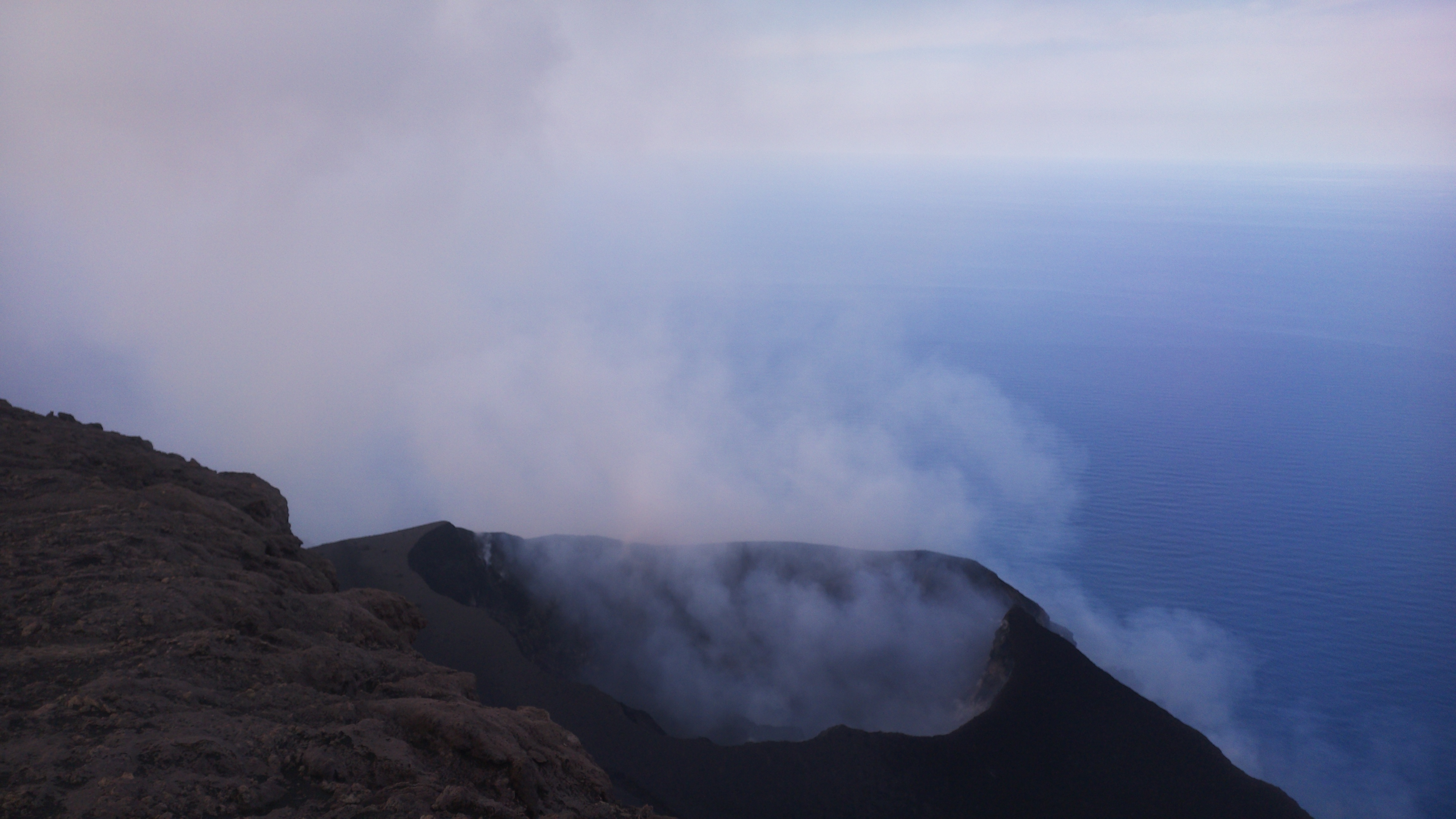}&
			\includegraphics[width=0.15\textwidth]{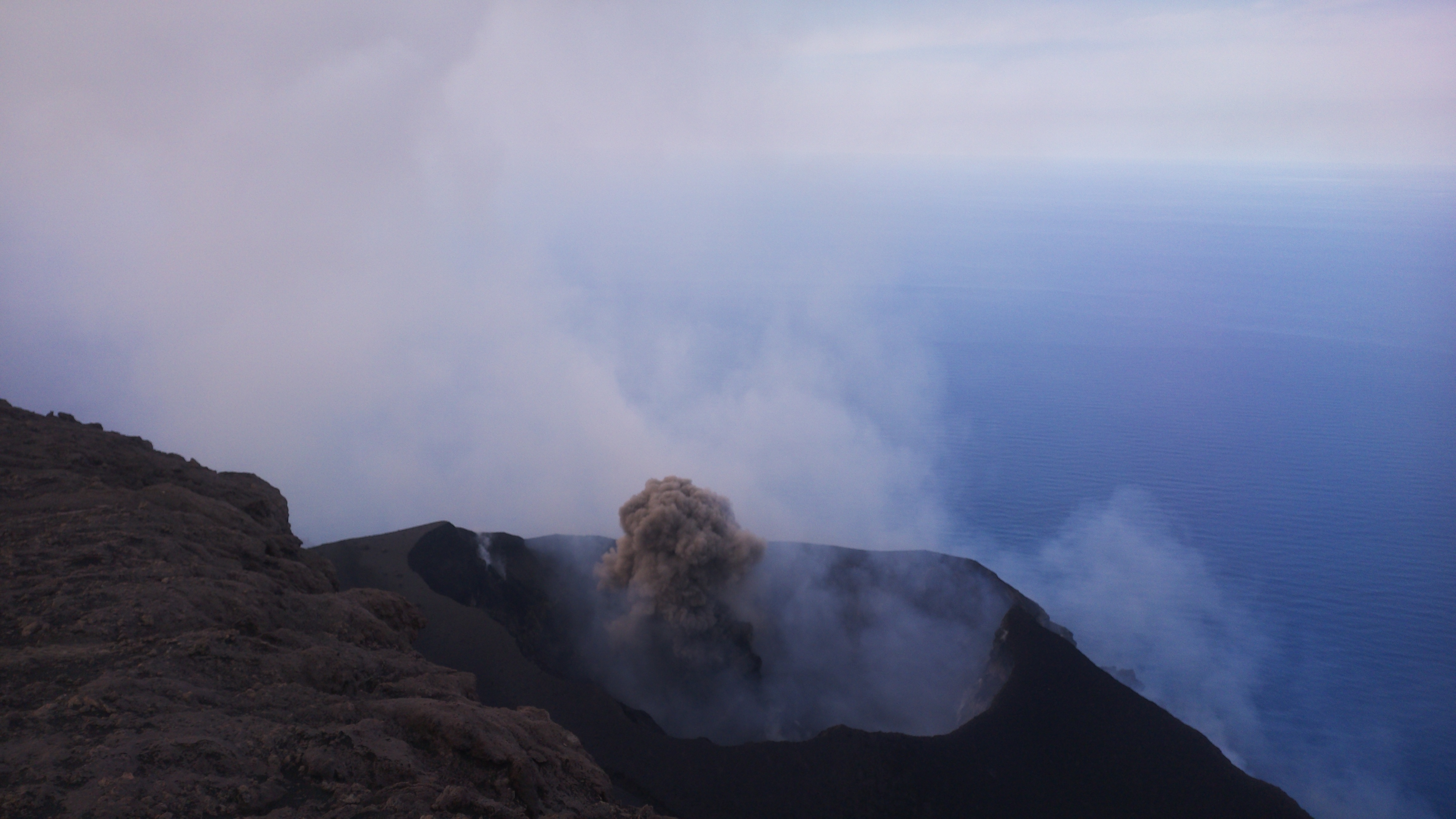}&
			\includegraphics[width=0.15\textwidth]{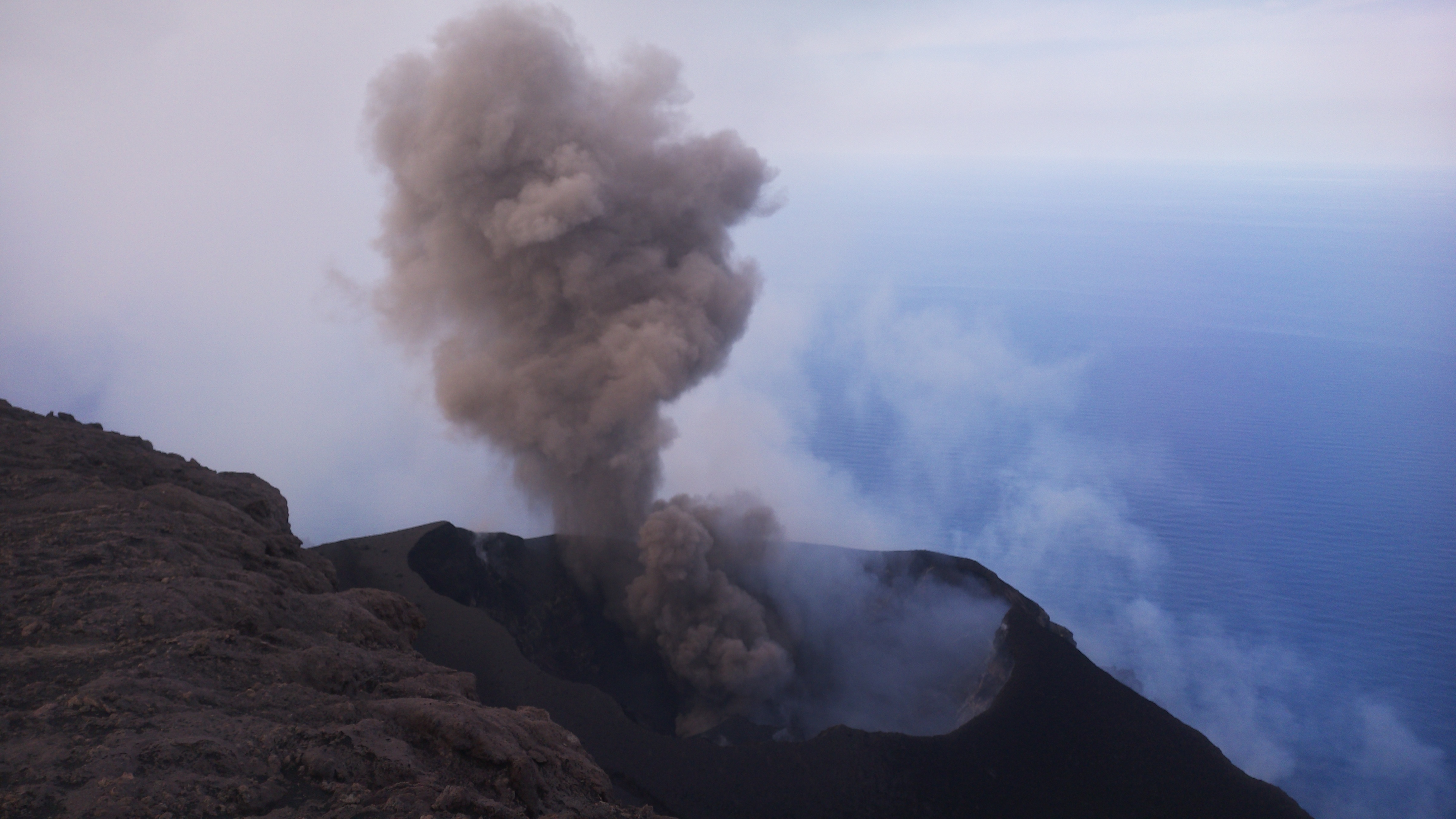}&
			\includegraphics[width=0.15\textwidth]{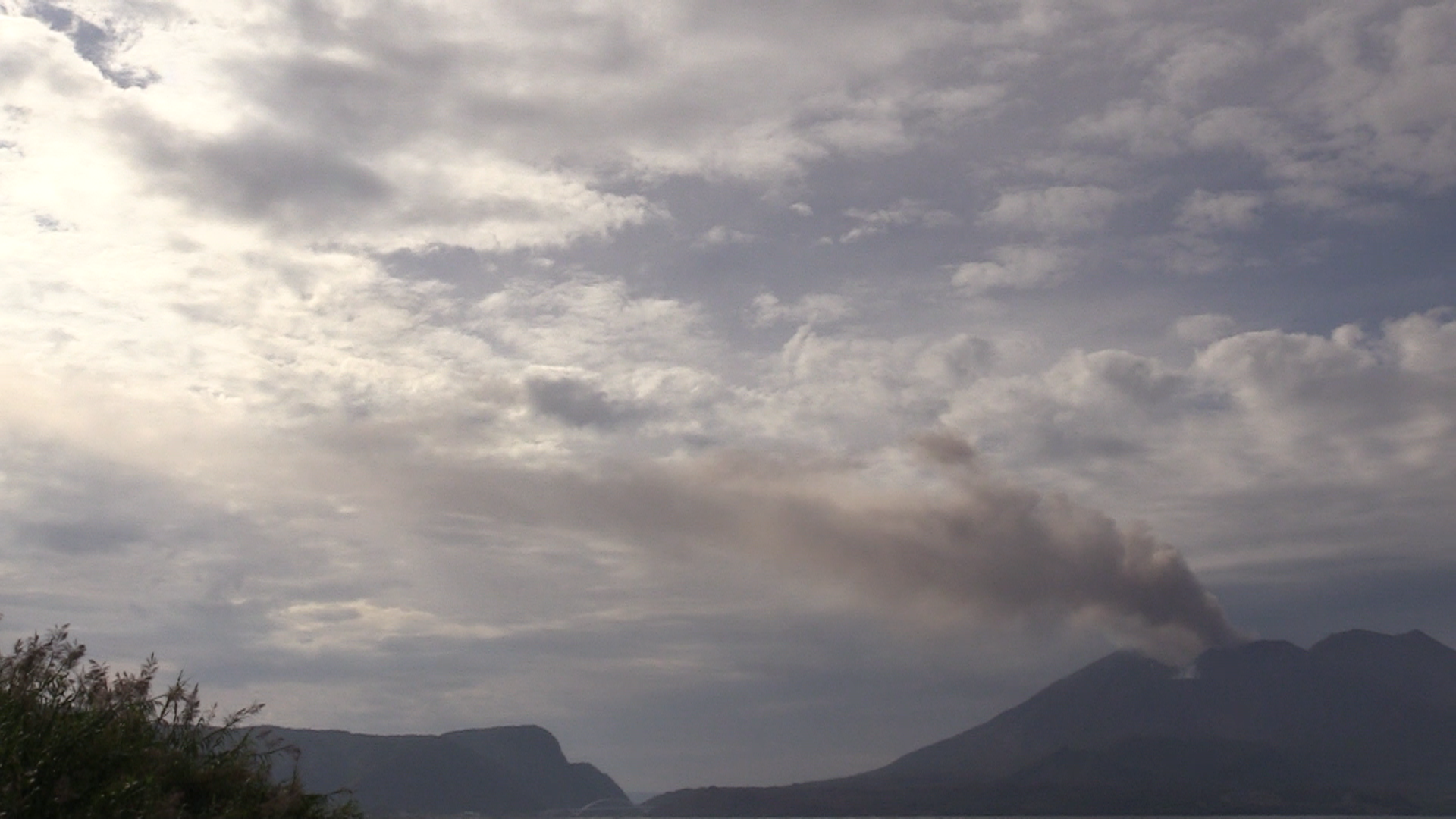}&
			\includegraphics[width=0.15\textwidth]{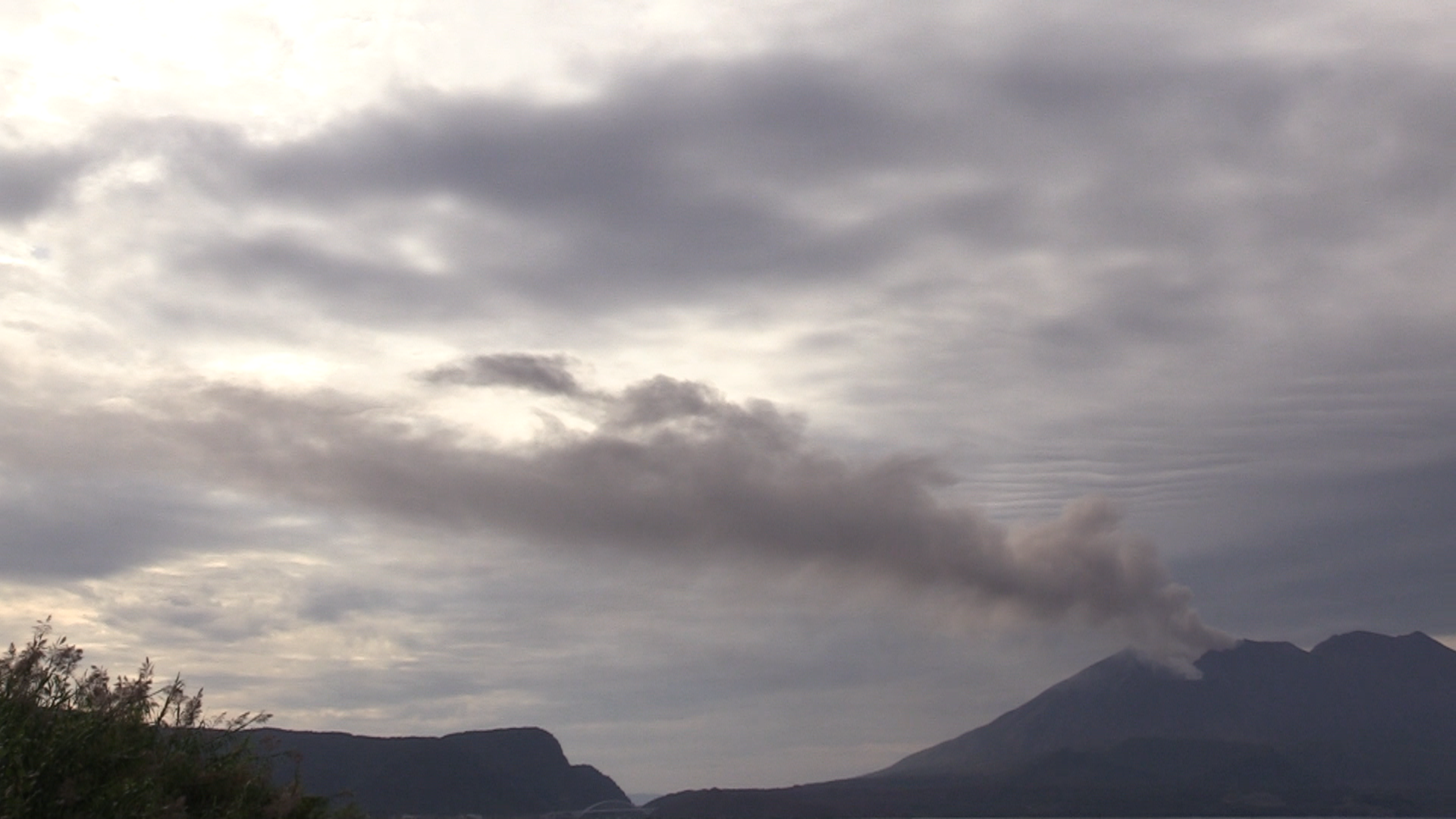}&
			\includegraphics[width=0.15\textwidth]{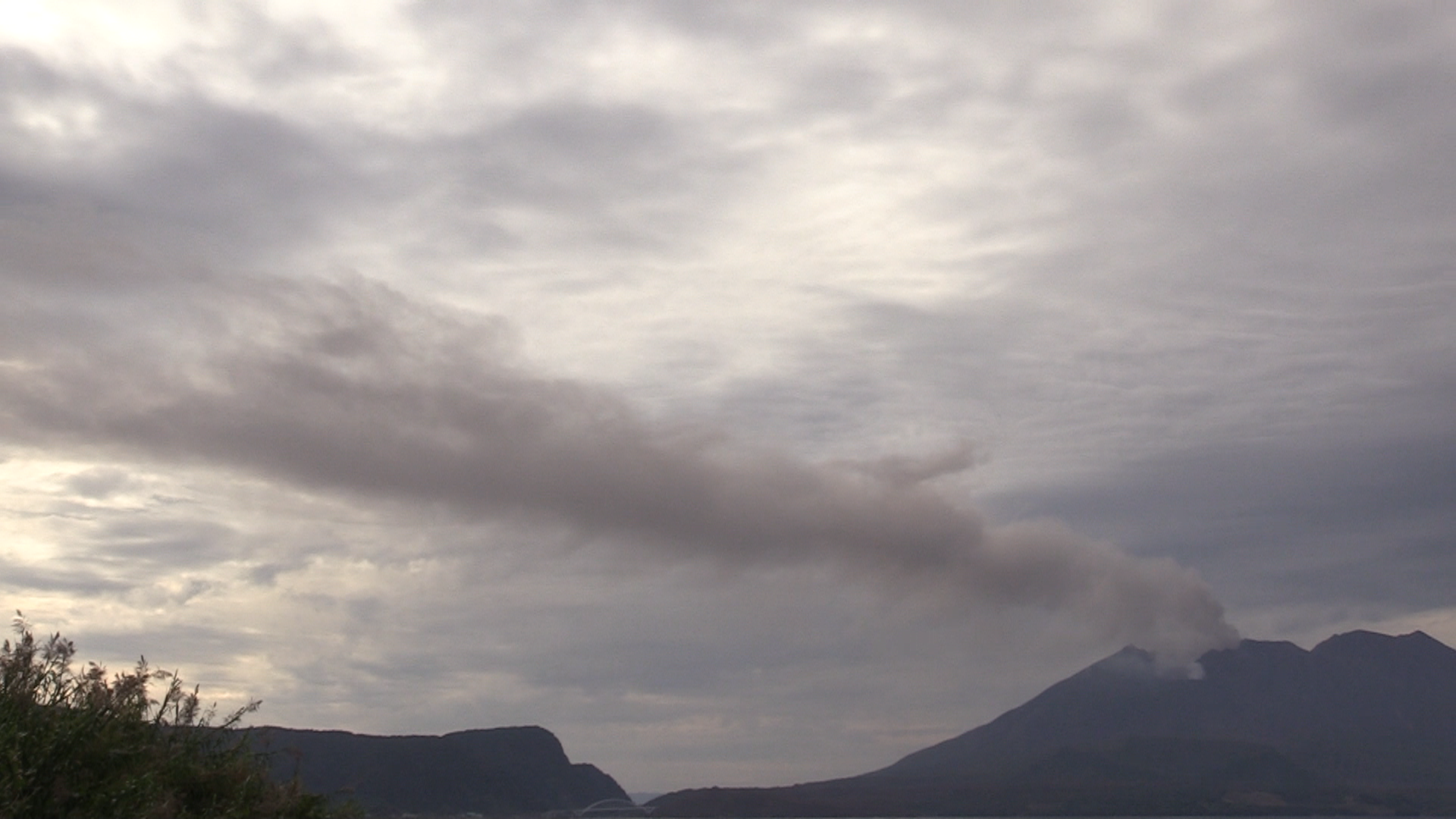}\\
		\end{tabular}
	\end{adjustbox}
    \vspace{-0.5cm}
	\caption{Dataset samples presenting three keyframes of four distinct sequences.}
	\label{fig:datasetgrid}
\end{figure}

The dataset includes images of volcanic ash plumes and clouds acquired from ground-based camera systems during field campaigns and monitoring explosive eruptions. It consists of 47 manually segmented time-series sequences totaling 759 frames, with a mean sequence length of 16 frames and a mean spatial resolution of $2996\times1685$ pixels. The sequences span a wide range of conditions, including varying viewing angles, illumination, occlusions, and weather, as well as diverse plume dynamics from early formation (rising plume state) to fully developed structures (cloud dispersal state). Although modest in size, the dataset is representative of the data scarcity typically encountered in volcanological field conditions, with annotations carefully designed to capture the physically meaningful extent and morphological characteristics of the plumes. The dataset is made publicly available alongside a re-annotated version produced by volcanologist guidance within the \emph{GuidedSAM-Plume} framework.  Figure \ref{fig:datasetgrid} presents a representative grid of samples from four video sequences included in the dataset.

\section{Results} \label{results}

\subsection{Comparison with fully automatic methods}

\subsubsection{Quantitative analysis}
\begin{table}[tbp]
	\caption{Segmentation performance metrics (\%) for volcanic plume segmentation, comparing Wilkes et al.~\cite{Wilkes2022}, VigiaPlume-Net~\cite{giffard2025vigia}, and SAM2\cite{Ravi2024}.}
		
	\begin{adjustbox}{width=\columnwidth}
		\begin{tabular}{|c||c|c|c|c|c|c|c|c|c|c|c|}
			\hline
			Method & $A$ & $P$ &  $R$ &	$D$ &	$J$	&$F$ &$J\&F$ &$\mathit{AR}{\epsilon}$&$\mathit{CVX}{\epsilon}$ & $\mathit{SLD}{\epsilon}$ \\
			\hline
			$W_{DenseNet121+UNet}$ \cite{Wilkes2022} &97.57&55.56&67.32&50.64&42.43&2.80&22.61&67.60&78.91&78.89\\
			\hline
			$W_{Inceptionv3+UNet}$  \cite{Wilkes2022} &97.88&65.59&63.79&52.53&45.41&4.39&24.90&65.59&74.63&76.74\\
			\hline
			$W_{EfficientNetB0+FPN}$ \cite{Wilkes2022}&96.72&58.45&59.73&42.37&36.07&3.65&19.86&61.15&71.96&74.54\\ 
			\hline
			VIGIA-PlumeNet \cite{giffard2025vigia}   &\textbf{\textit{98.35}}&\textbf{93.69}&42.57&43.15&37.85&6.33&22.09&47.45&49.65&52.22\\
			\hline
			SAM2 (1 point prompt)\cite{Ravi2024}&96.84&85.88&\textbf{\textit{83.25}}&\textbf{\textit{78.68}}&\textbf{\textit{71.24}}&\textbf{\textit{7.23}}&\textbf{\textit{39.23}}&\textbf{\textit{84.76}}&\textbf{\textit{87.78}}&\textbf{\textit{91.79}}\\
			\hline
			\hline
			\emph{GuidedSAM-Plume} Workflow&\textbf{99.54}&93.34&\textbf{90.35}&\textbf{91.13}&\textbf{84.69}&\textbf{12.89}&\textbf{48.79}&\textbf{92.02}&\textbf{90.93}&\textbf{93.01}\\
			\hline
		\end{tabular}
	\end{adjustbox}

	\label{tab:inference}
\end{table}

The \emph{GuidedSAM-Plume} workflow  was compared against existing dedicated fully automatic solutions for volcanic plume and cloud segmentation as well as a single-point SAM2 baseline. Following the protocol described in Section~\ref{protocol}, inference was performed for the models proposed by Wilkes et al.~\cite{Wilkes2022} and Giffard-Roisin et al.~\cite{giffard2025vigia}, as well as for SAM2~\cite{Ravi2024} with a single randomly selected positive prompt (drawn from \emph{within} the ground truth mask). Results are reported in Table~\ref{tab:inference}. While VigiaPlume-Net achieves the highest accuracy and precision, it underperforms notably on recall and volcanological metrics. Even without any refinement, SAM2 considerably outperforms existing solutions, demonstrating the value of exploiting its zero-shot capabilities for volcanic plume segmentation. An overall lower performance is observed on the boundary-sensitive \emph{F} measure across all methods, though SAM2 also achieves higher scores on this metric and results further show that it can be improved on a case-by-case basis. In this regard, final results obtained with the full expert-guided \emph{GuidedSAM-Plume} framework are also reported, showing considerable additional improvements over the fully automated methods and confirming that targeted expert refinement further enhances segmentation performance of SAM2. The observed performance gains substantiate that the additional annotation effort required by the framework is well justified, supporting its adoption for quantitative volcanological studies.

\subsubsection{Qualitative analysis of challenging scenes}
\begin{figure}[tbp]
\begin{adjustbox}{width=\textwidth}
	  \begin{tabular}{ccc@{\hspace{0.2cm}}cc}
	    \multicolumn{3}{c}{{\scriptsize{\textit{GuidedSAM-Plume}}}} &{\scriptsize{ W\textsubscript{DenseNet121+UNet}}~\cite{Wilkes2022}}
	     & {\scriptsize{VigiaPlume-Net~\cite{giffard2025vigia}}}\\
	    \includegraphics[width=0.19\textwidth]{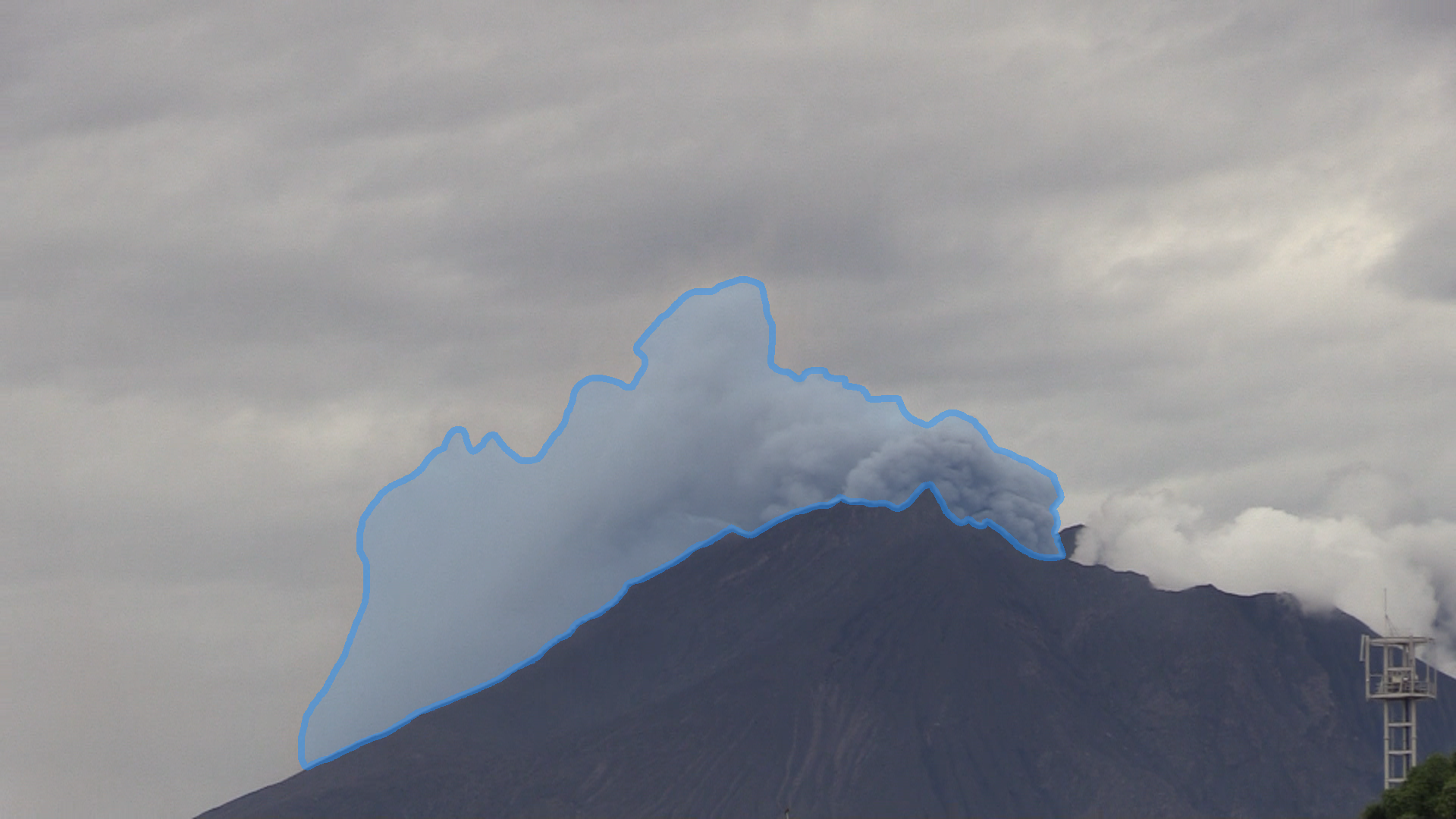}&
	    \includegraphics[width=0.19\textwidth]{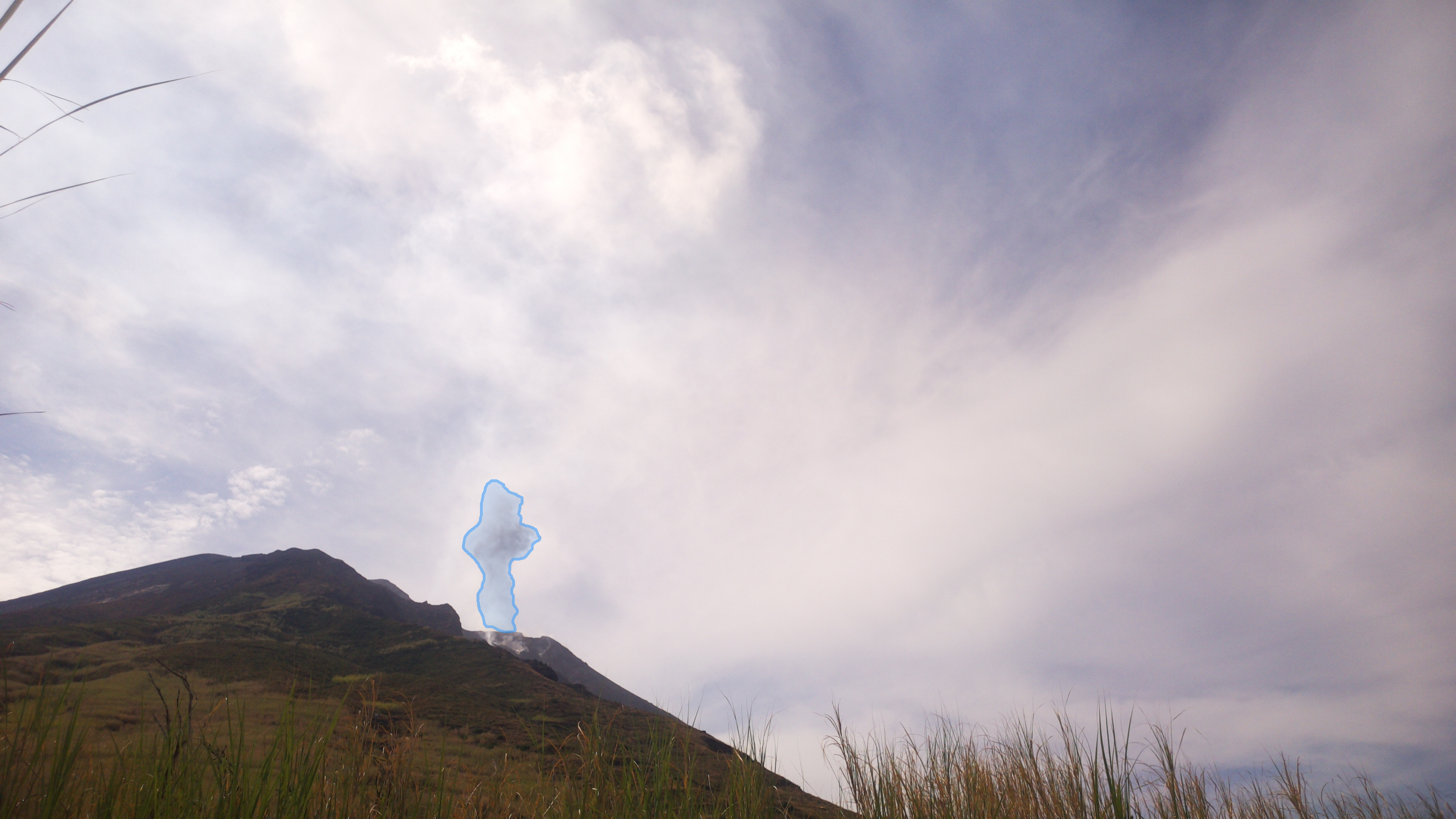}&
	    \includegraphics[width=0.19\textwidth]{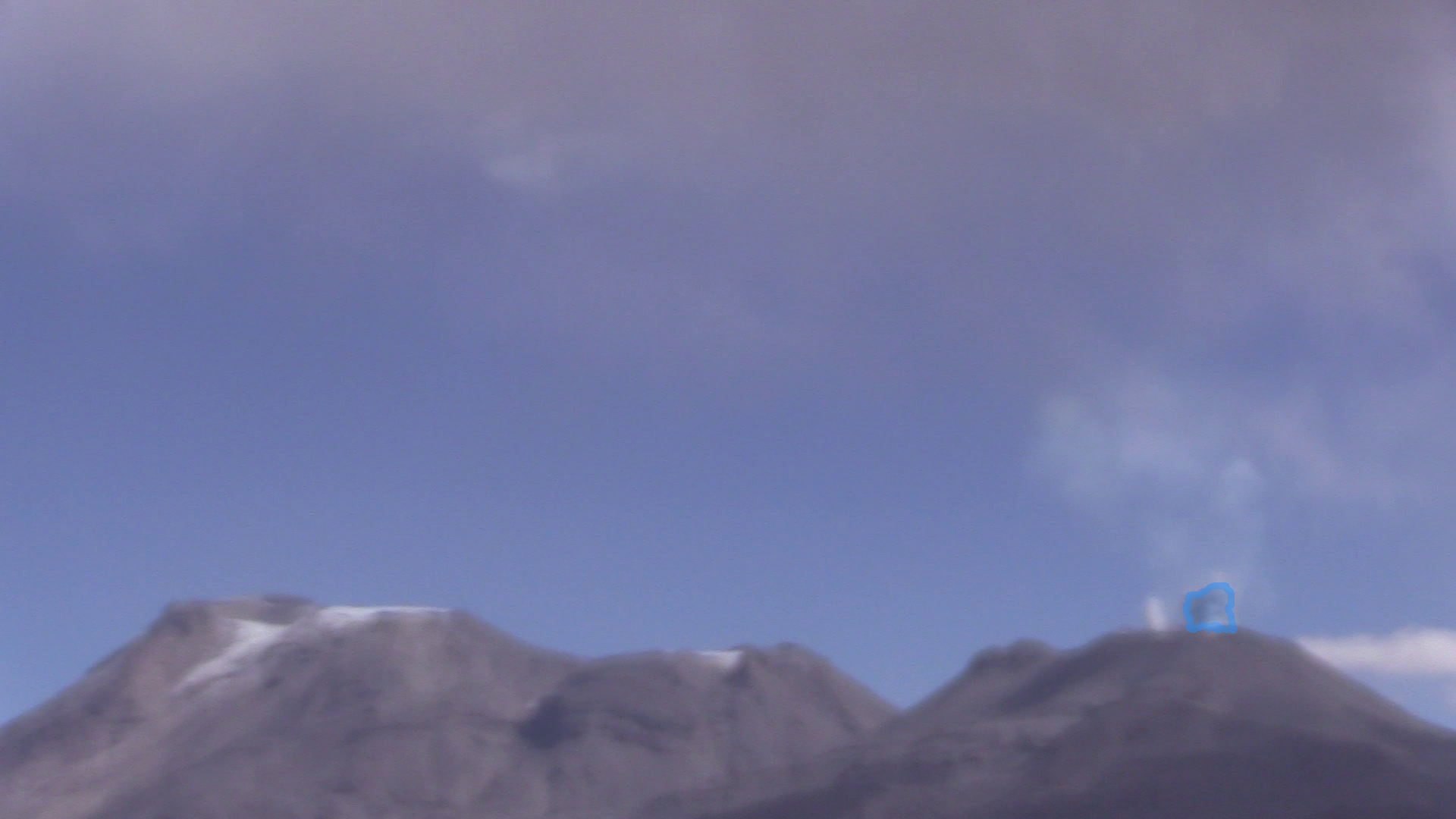}&
	    \includegraphics[width=0.19\textwidth]{figures/gtStromboli.png}&
	    \includegraphics[width=0.19\textwidth]{figures/gtSabancayaMask.png}\\
	    \includegraphics[width=0.19\textwidth]{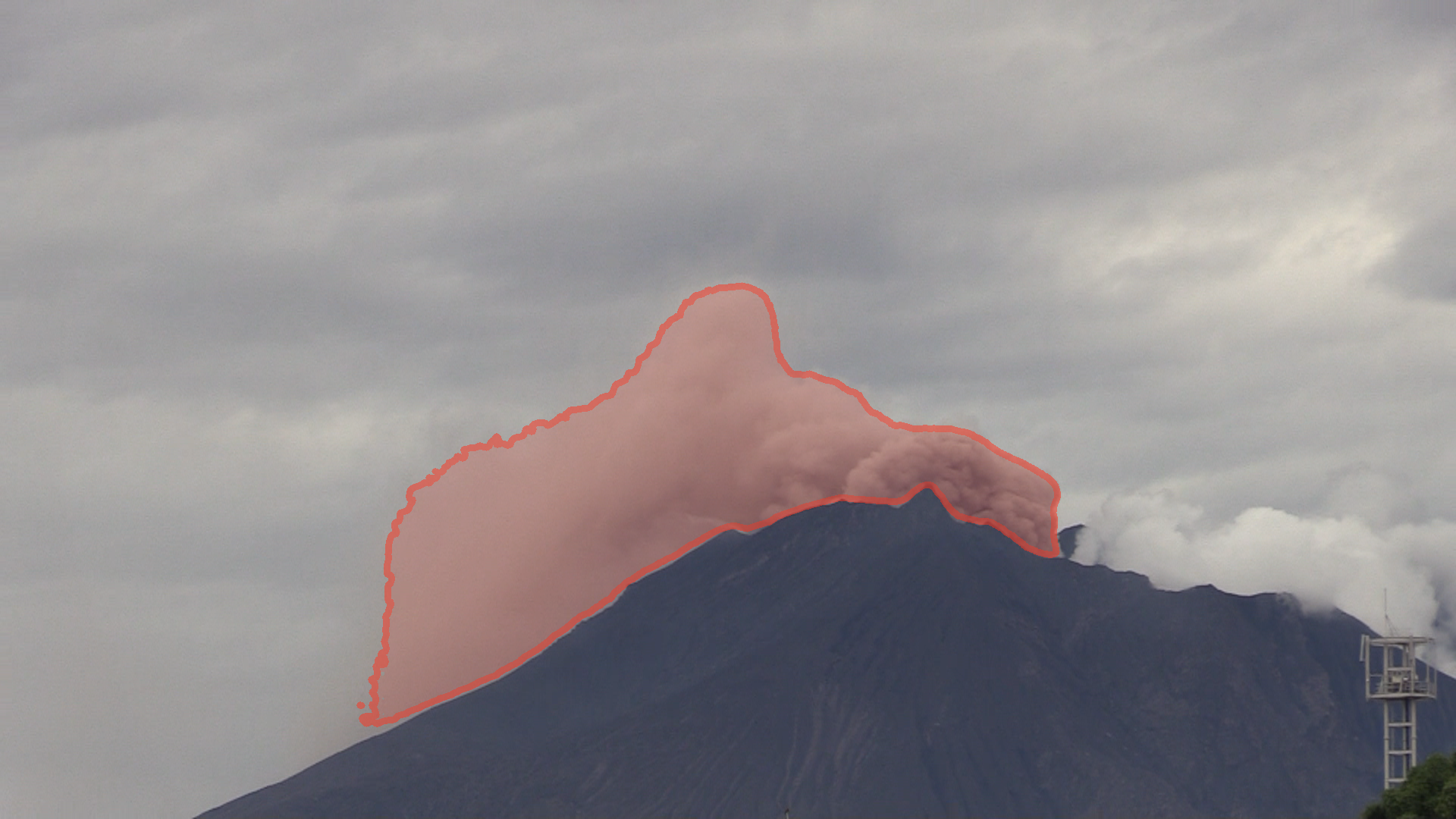}&
	    \includegraphics[width=0.19\textwidth]{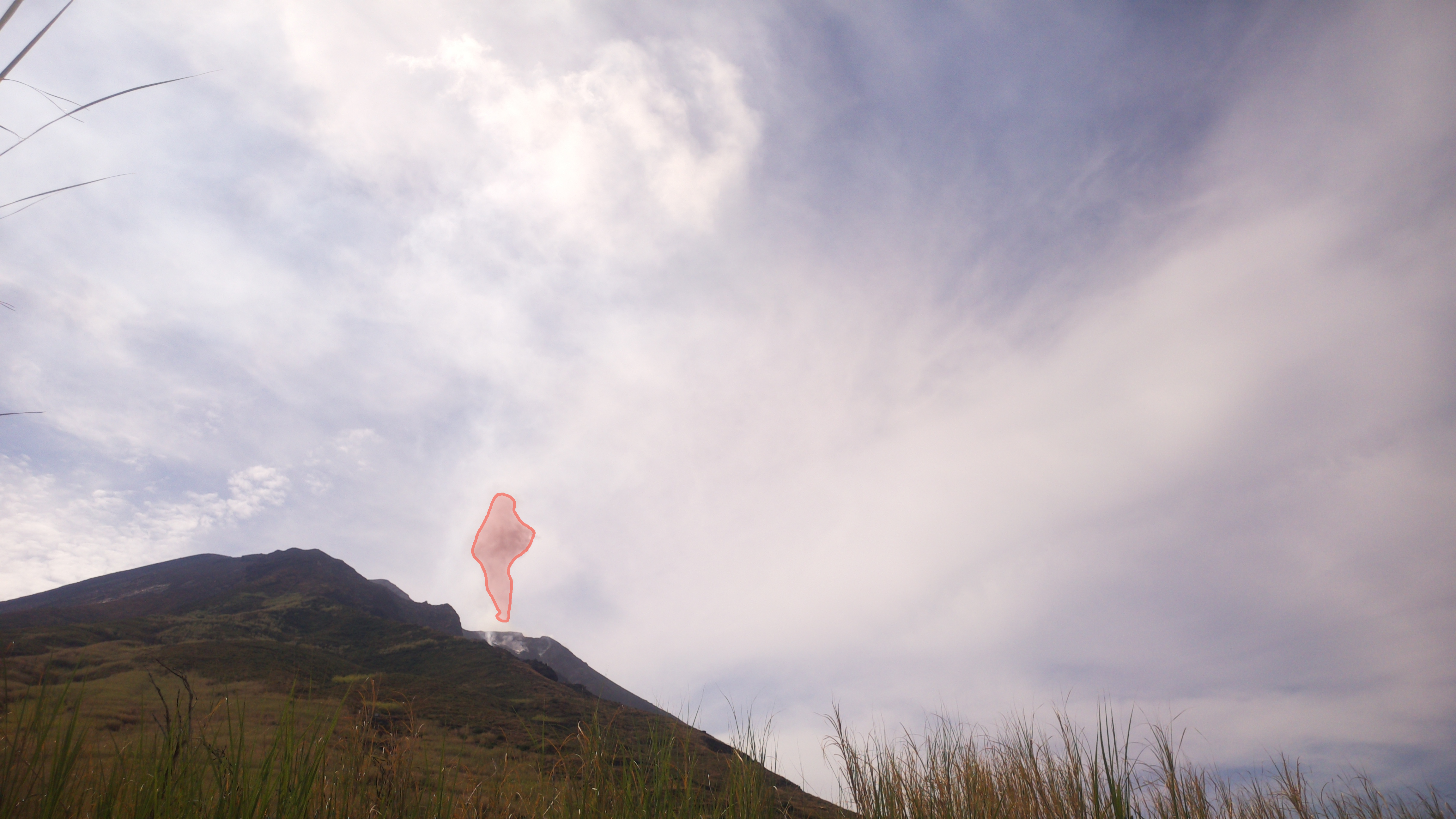}&
	    \includegraphics[width=0.19\textwidth]{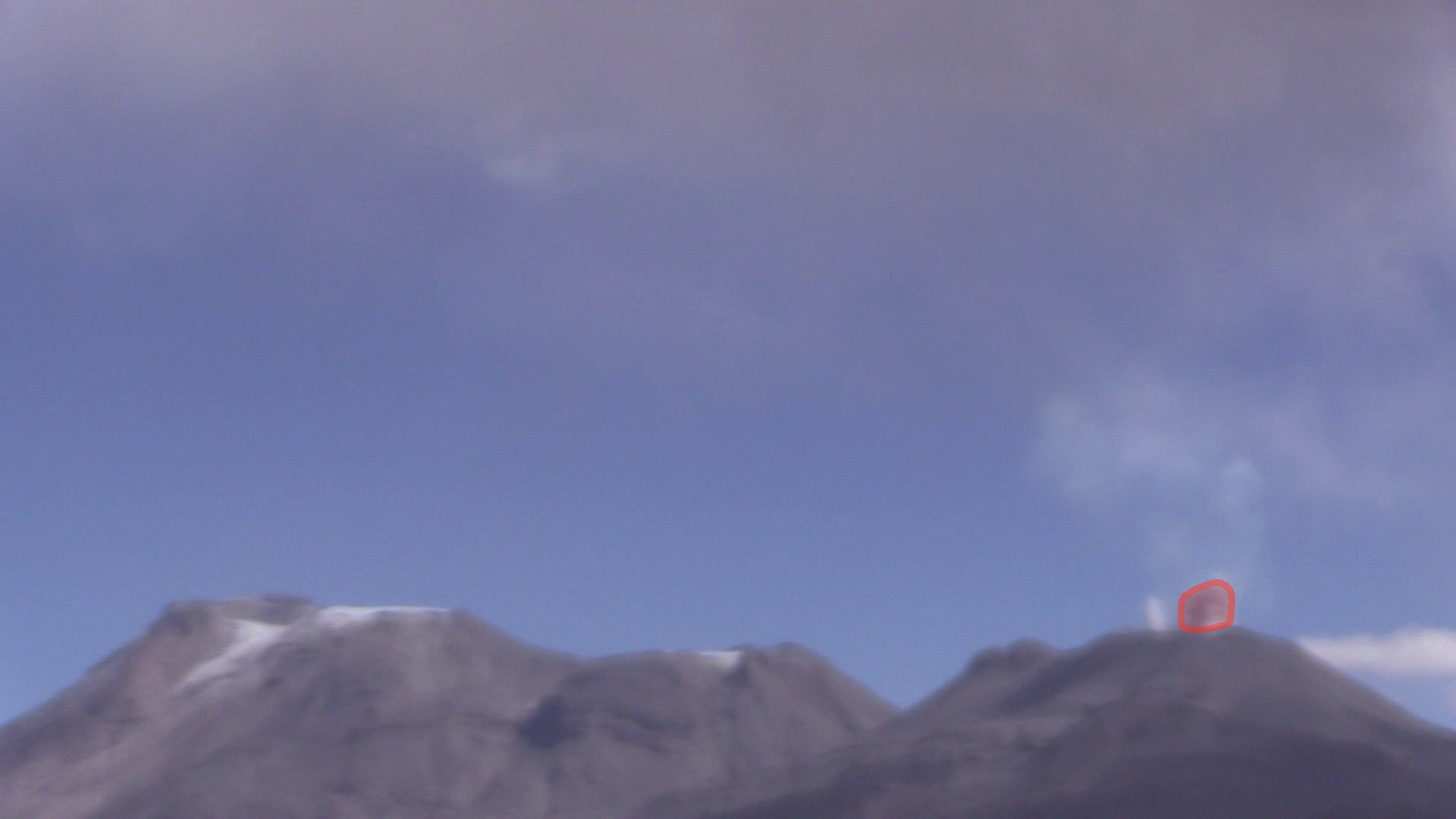}&
	    \includegraphics[width=0.19\textwidth]{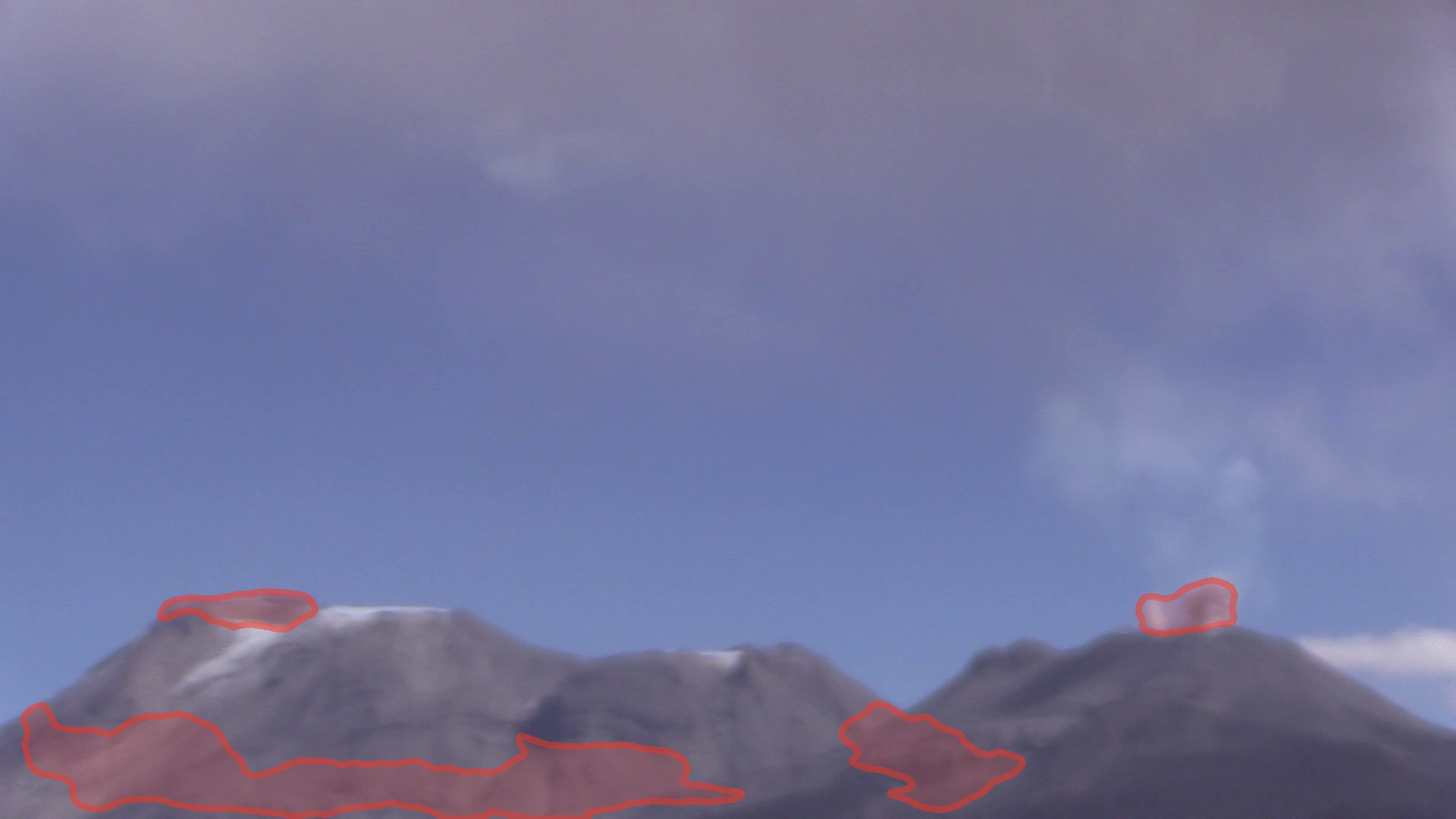}&
	    \includegraphics[width=0.19\textwidth]{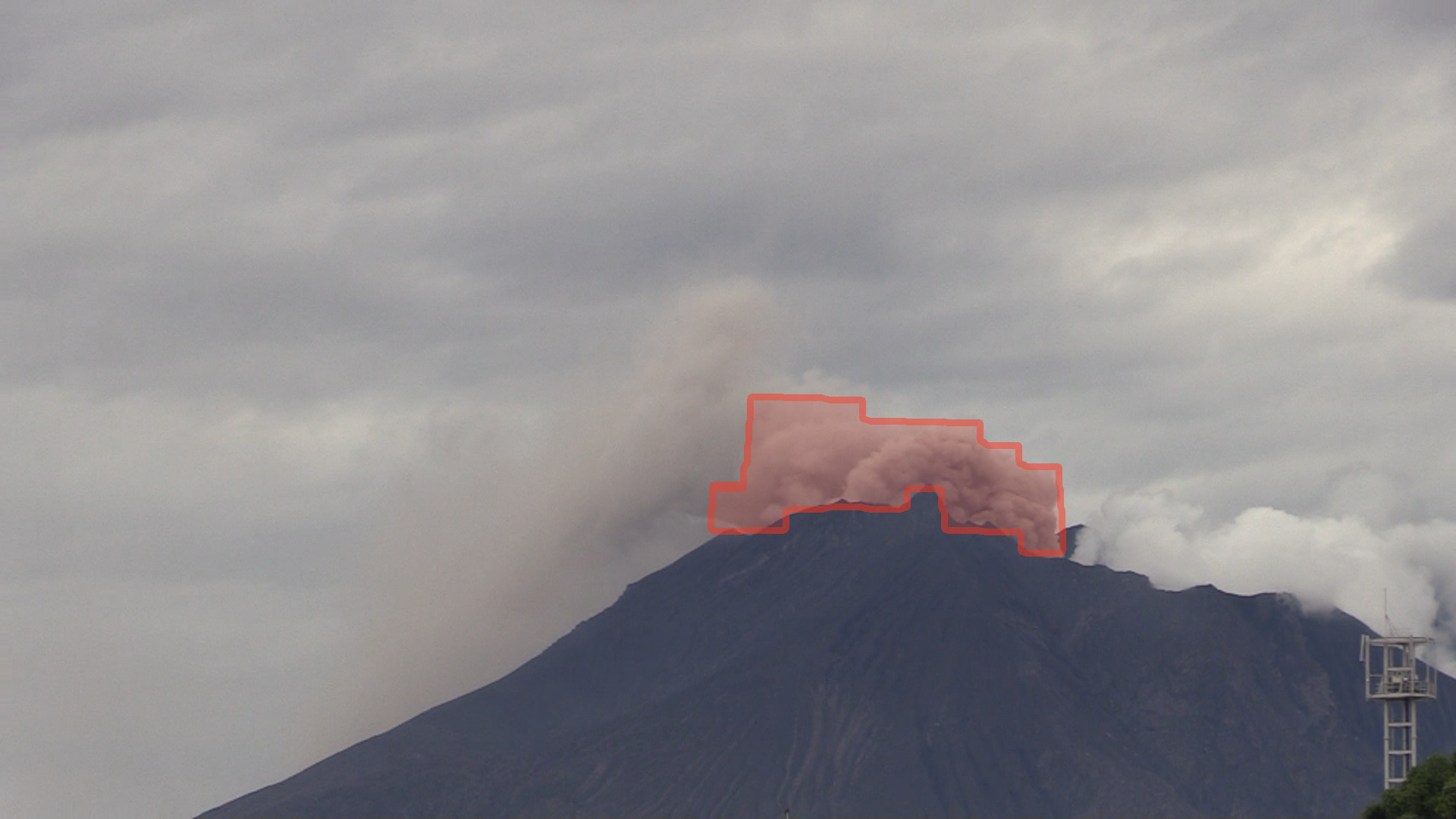}\\
	  \end{tabular}
\end{adjustbox}
\vspace{-0.5cm}
	\caption{Qualitative comparison of segmentation outputs across models on the most challenging scenes of the dataset.  Ground-truth annotations are shown in blue and outputs in red.}
	\label{fig:qualitative}
\end{figure}
Visual inspection of segmentations, particularly in the most challenging scenes, is necessary to better interpret the relative effectiveness of the compared models and the significance of observed metric discrepancies. Figure~\ref{fig:qualitative} presents the scenes yielding the worst segmentation results under the \emph{GuidedSAM-Plume} framework. In these cases, degraded performance is primarily attributable to highly diffuse boundaries, heterogeneous opacity, small plume extents amplifying relative errors, or out-of-focus imagery. It is worth noting that under such conditions, plume delineation may vary between human annotators and therefore does not necessarily reflect a strict segmentation error. Given the acceptable produced segmentation even in these conditions, a further advantage of the proposed framework lies in its ability to standardize ambiguous delineation decisions and reduce inter-annotator variability.
For the same scenes, the models of Wilkes et al. exhibit a tendency to generate spurious plume detections on background regions. VigiaPlume-Net systematically produces staircase boundary artifacts that fail to preserve plume geometric characteristics, limiting its suitability for downstream quantitative analysis. 

\subsection{Maximal expected performance}\label{best_achievalbe}
\begin{table}[tbp]
	\centering
	\caption {Best achievable performance and minimal error for morphological descriptors (\%). Poor performances are highlighted in bold.}
	\begin{adjustbox}{width=0.8\columnwidth}
		\begin{tabular}{|cccccccccc|}
			\hline
			\multicolumn{10}{|c|}{Segmentation metrics}\\
			\hline
			$A$ & $P$ &  $R$ &	$D$ &	$J$	&$F$ &$J\&F$ &$\mathit{AR}{\epsilon}$&$\mathit{CVX}{\epsilon}$ & $\mathit{SLD}{\epsilon}$\\
			\hline
			99.97&99.09&97.64&98.05&96.77&\textbf{38.08}&\textbf{67.43}&98.43&\textbf{93.55}&97.4\\
			
			\hline
		\end{tabular}
	\end{adjustbox}
	\begin{adjustbox}{width=\columnwidth}
		\begin{tabular}{|cccccc||cccccc|}
			\hline
			\multicolumn{12}{|c|}{Morphology parameters}\\
			\hline
			\multicolumn{6}{|c||}{mean \textit{\textbf{absolute}} relative errors}&\multicolumn{6}{c|}{mean \textbf{\textit{signed}} relative errors}\\
			\hline
			$\mathit{CP}_\mathrm{M}$&$\mathit{CP}_\mathrm{CH}$&$\mathit{SA}_\mathrm{M}$&$\mathit{SA}_\mathrm{CH}$&$\mathit{FD}_{\max}$&$\mathit{FD}_{\min}$
			&$\mathit{CP}_\mathrm{M}$&$\mathit{CP}_\mathrm{CH}$&$\mathit{SA}_\mathrm{M}$&$\mathit{SA}_\mathrm{CH}$&$\mathit{FD}_\mathrm{max}$&$\mathit{FD}_\mathrm{min}$\\
			\hline
			\textbf{7.44}&2.06&1.45&3.40&2.04&2.04&\textbf{-7.44}&-2.06&-1.37&-3.40&-2.03&-2.03\\
			\hline
		\end{tabular}
	\end{adjustbox}
	\label{tab:minloss}
\end{table}
\noindent To better quantify the intrinsic limitations of the proposed framework, downscaling and reupscaling manipulations where performed on ground truths as described in Section~\ref{protocol}. The resulting performance over the metrics between the original ground truth and the downscaled then re-upscaled ones are reported in Table~\ref{tab:minloss}. The most critical errors to consider are those affecting the volcanological descriptors, as these directly reflect the impact on downstream analyses. To better characterize degradation trends, both absolute and signed relative errors with respect to the ground truth are reported for the underlying morphological descriptors from which these metrics are derived. For most metrics, the loss is negligible but large discrepancies are observed in metrics that rely on edge measurements, i.e. $F$, $J\&F$ and $\mathit{CVX}{\epsilon}$. This is expected, as boundaries are highly sensitive to scaling processes. The contour perimeter, convex hull perimeter and area exhibit identical mean relative and absolute errors, indicating a unidirectional and consistently negative bias. Notably, the magnitude of the perimeter bias is consistent with loss on convexity exceeding 6\%, while errors in solidity are primarily driven by a slight underestimation of the convex hull area. These observations reveal an unavoidable smoothing effect intrinsic to the SAM2 architecture. Thus established lower bound on performance provides a reference baseline for the evaluation of empirical results.

\subsection{Workflow establishment}

\subsubsection{Prompts efficiency} \label{prompting_result}
\begin{table}[tbp]
	\centering
	\caption{Comparison of segmentation results across different prompt configurations, combining foreground points (FP), background points (BP), and bounding boxes (BB). When applicable, the \emph{center} corresponds to the brightest point in the distance map.}
	\begin{adjustbox}{width=\columnwidth}
		\begin{tabular}{|c|ccc||c|c|c|c|c|c|c|c|c|c|}
			\cline{2-14}
			\multicolumn{1}{c|}{}&FP &BP &BB& $A$ & $P$ &  $R$ &	$D$ &	$J$	&$F$ &$J\&F$ &$\mathit{AR}{\epsilon}$&$\mathit{CVX}{\epsilon}$ & $\mathit{SLD}{\epsilon}$ \\
			\cline{2-14}
			\noalign{\vskip 2pt}  
			\hline
			\multirow{3}{*}{\makecell{Single\\prompt}}
			&0&0&True&\textbf{99.3}&\textbf{89.0}&\textbf{93.4}&\textbf{89.9}&\textbf{83.4}&\textbf{7.8}&\textbf{45.6}&\textbf{91.8}&\textbf{89.3}&\textbf{94.2}\\
			&1(center)&0&False&97.5&87.2&84.1&80.6&73.0&7.4&40.2&85.2&87.8&91.8\\
			&1&0&False&96.8&85.9&83.2&78.7&71.2&7.2&39.2&84.8&87.8&91.8\\
			\hline\hline
			\multirow{4}{*}{\makecell{No \\bounding\\box}}
			&5&0&False&98.1&87.0&\textbf{89.9}&\textbf{85.0}&\textbf{78.2}&7.5&\textbf{42.8}&\textbf{89.7}&\textbf{\textit{88.3}}&93.0\\
			&4&0&False&97.7&86.6&89.5&84.3&77.4&7.5&42.5&89.2&\textit{\textbf{88.3}}&92.9\\
			&4&1&False&\textbf{\textit{98.4}}&87.0&88.7&84.3&77.1&7.4&42.3&89.2&\textbf{\textit{88.3}}&\textbf{93.1}\\
			&3&1&False&\textbf{\textit{98.4}}&\textbf{87.5}&88.0&84.2&77.1&7.4&42.3&89.0&\textbf{\textit{88.3}}&93.0\\
			\hline\hline
			\multirow{4}{*}{\makecell{Best\\results}}
			&0&1&True&99.3&\textbf{89.2}&93.1&90.0&83.4&\textbf{\textit{7.8}}&\textbf{\textit{45.6}}&91.8&89.3&94.2\\
			&0&0&True&99.3&89.0&93.4&89.9&83.4&\textbf{\textit{7.8}}&\textbf{\textit{45.6}}&91.8&89.3&94.2\\
			&1&2&True&\textbf{99.5}&88.9&\textbf{\textit{93.7}}&\textbf{\textit{90.1}}&\textbf{\textit{83.6}}&7.7&\textbf{\textit{45.6}}&\textbf{92.0}&\textbf{\textit{89.5}}&\textbf{\textit{94.4}}\\
			&1&3&True&99.4&88.8&\textbf{\textit{93.7}}&\textbf{\textit{90.1}}&\textbf{\textit{83.6}}&7.7&\textbf{\textit{45.6}}&91.9&\textbf{\textit{89.5}}&\textbf{\textit{94.4}}\\
			\hline
		\end{tabular}
	\end{adjustbox}
	\label{tab:prompt}
\end{table}
Representative results of SAM2 across various prompt combinations are
reported in Table~\ref{tab:prompt}. When using a single prompt, a
bounding box consistently outperforms a single positive point. Note that negative points are not meaningful in this single prompt setting. No
significant performance difference was observed between a randomly
selected point and one located at the center of the distance
map. These findings align with observations reported in the original
SAM study \cite{Kirillov2023} and underscore the value of an
expert-guided strategy, where prompts are informed by domain knowledge
rather than predetermined spatial patterns. When no bounding box is
included, overall segmentation performance declines markedly, even
when four to five point prompts are provided, while precision remains
relatively stable. This suggests that the segmentation workflow should
be designed such that operators initiate the process with a bounding
box, which provides a strong spatial prior. Furthermore, as shown in the lower section of Table~\ref{tab:prompt}, supplementing the bounding box with randomly placed points yields no significant improvement; the best results are consistently obtained with a
bounding box complemented by a small number of selectively placed
points, with a preference for background points. Unnecessary or
excessive point prompts may in fact be counterproductive. Point
prompts should therefore be incorporated iteratively and selectively,
targeting regions where refinement is genuinely needed rather than
maximizing the number of prompts. These results emphasize that an
expert-guided approach, initialized with a bounding box and
complemented by selective point prompts, provides the most efficient
and reliable strategy for SAM2-based segmentation of complex natural
structures.

\subsubsection{Video propagation assessment}
\begin{table}[tbp]
	\centering
	\caption{Video propagation performance depending on the position of the prompted frame in the sequence (\%). \emph{First} refers to the first frame that is not purely background in the sequence.}
	\begin{adjustbox}{width=\columnwidth}
		\begin{tabular}{|c||c|c|c|c|c|c|c|c|c|c|}
			\hline
			Prompted keyframes& $A$ & $P$ &  $R$ &	$D$ &	$J$	&$F$ &$J\&F$ &$\mathit{AR}{\epsilon}$&$\mathit{CVX}{\epsilon}$ & $\mathit{SLD}{\epsilon}$ \\
			\hline
			First&99.30&\textbf{95.53}&74.85&77.84&71.22&12.52&41.87&79.08&79.65&83.34\\
			\hline
			Middle&\textbf{\textit{99.53}}&91.55&\textbf{\textit{92.03}}&\textbf{\textit{89.72}}&\textbf{\textit{84.38}}&\textbf{\textit{14.40}}&\textbf{\textit{49.39}}&\textbf{\textit{91.14}}&86.26&\textbf{\textit{91.45}}\\
			\hline
			Last&99.31&89.86&90.68&87.96&82.40&14.32&48.36&89.95&\textbf{\textit{86.57}}&91.21\\
			\hline
			\hline
			Middle + First + Last&\textbf{99.63}&92.97&\textbf{94.26}&\textbf{92.87}&\textbf{87.91}&\textbf{17.74}&\textbf{52.82}&\textbf{93.91}&\textbf{88.45}&\textbf{93.76}\\
			\hline
			
		\end{tabular}
	\end{adjustbox}
	\label{tab:videopropagresults}
\end{table}

We assessed the quality of mask propagation depending on the annotated keyframes. In Table~\ref{tab:videopropagresults}, we present the results of the different propagation performances as a function of the position of the prompted frame, averaging the results over the remaining frames in the sequence. Propagation was initialized from ground-truth masks as input, reflecting an interactive setting in which an expert can readily refine the input mask prior to propagation. Results indicate that propagation performance remains relatively stable across video sequences. Notably, the boundary F-score exceeds those previously observed for static image segmentation, suggesting that temporal propagation does not degrade contour accuracy compared to what can be achieved on a single frame. Initializing propagation from the beginning or end of the sequence yields comparable trends, though performance is lower when starting from the first frame, likely due to the limited spatial extent of the plume in early sequence stages. A slight advantage is observed when initializing from a middle frame, where the plume is already sufficiently developed and fewer frames are subject to error propagation. Last line in the Table~\ref{tab:videopropagresults} show that performance is further improved when propagation is initialized from the middle frame and subsequent keyframes are added. Overall, SAM2 video propagation proves reliable and well-suited to real-world conditions. For practical use, initializing propagation from a middle frame is recommended, with targeted expert refinement applied as needed.

\subsubsection{Impact on volcanological applications} \label{volcanoimpact}

\begin{figure}[tbp]
	\centering
	\includegraphics[width=\linewidth]{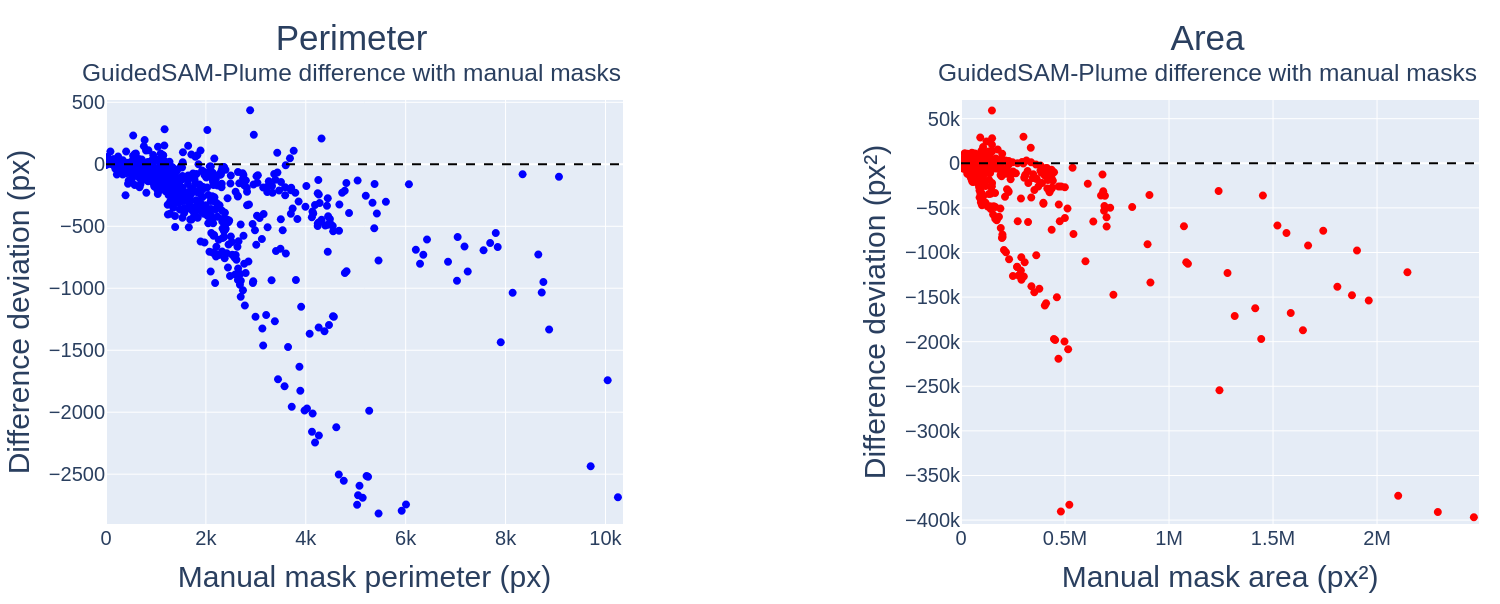}
        \vspace{-0.5cm}
	\caption{Deviation of perimeter (blue) and area (red) estimates derived from \emph{GuidedSAM-Plume} masks with respect to manually annotated reference masks, as a function of the reference values. Each marker corresponds to an individual annotated frame.}
	\label{fig:perimeter_area_deviations}
\end{figure}

We investigated the impact of masks produced by the
\emph{GuidedSAM-Plume} framework on downstream volcanological
analyses. As previously reported in Table~\ref{tab:inference}, all
metrics derived from volcanological parameters exceed 90\% with the
complete workflow, indicating reliable performance. To further
characterize this agreement, we examine two  geometric
descriptors: plume perimeter and area.  Their deviations to manually
produced reference masks, as a function of these quantities, are
presented in Figure~\ref{fig:perimeter_area_deviations}. A
predominantly negative deviation is observed for perimeter of masks
produced with the framework, reflecting a boundary smoothing effect
that becomes more pronounced and dispersed with increasing plume size.
A similar trend being observed for area, the boundary smoothing effect
is predominantly concave; in other words, mask boundaries mostly
manifest an inward contraction. These tendencies warrant careful
consideration when exploiting \emph{GuidedSAM-Plume} outputs for
downstream quantitative analyses. Nevertheless, the observed
deviations remain proportionally small, particularly for area
estimates which are less sensitive to contour errors. Moreover, such
deviations are arguably better controlled through a standardized
framework than the inter-annotator variability inherent to purely
manual annotation.

\section{Discussion and conclusion}

\textit{GuidedSAM-Plume} introduces a validated interactive segmentation framework for volcanic plume and cloud analysis, combining automated processing with expert-guided refinement of ambiguous regions. It offers a balance between expert intervention and processing efficiency compared to existing workflows, while maintaining strong agreement with ground-truth segmentation and preserving the reliability of derived volcanological measurements. In practice, expert annotation time was reduced from approximately 2 hours to 5–10 minutes per 10-frame sequence, varying with image resolution and scene complexity\footnote{Improvements observed on a workstation equipped with an NVIDIA GeForce RTX 4080 (16 GB vRAM) GPU.}. Moreover, deviations observed with manual masks in morphometric parameters that directly influence volcanological analyses remain controlled and predominantly systematic in direction, which strengthens the overall reliability.  Correcting for these systematic biases in downstream analyses represents a promising direction to further improve the accuracy of derived volcanological measurements. Further validation on second-order parameters, including real-scale plume dimensions and dynamic properties, would be valuable to more comprehensively confirm the robustness of the framework for operational  applications.

It should be noted that the present contribution is subject to several limitations. Refinement capabilities are inherently bounded by the design of SAM models, which were trained on a limited number of input points or short video sequences \cite{Kirillov2023, Ravi2024}. Moreover, although it improves upon fully automatic methods, it cannot dispense with expert intervention. The modest dataset size narrows the scope of quantitative evaluation, notably the relatively short size of temporal sequences. These limitations reflect the typical constraints inherent to volcanological studies. Nevertheless, wider adoption of the framework has the potential to facilitate the expansion of available data in this domain. In order to address the broader needs of the volcanological community, future development should integrate real-scale physical metric conversion, and incorporating multi-scale processing capabilities spanning from plume-level to particle-level characterization within a unified workflow.

Overall, the results highlight the potential of zero-shot foundation models to improve the reproducibility and efficiency of segmentation-based analysis for volcanology despite limited data availability.  The absence of requirements for heavy domain-specific retraining is expected to encourage adoption within the volcanological community. The proposed framework has already been effectively deployed under real-world settings by volcanologists to produce the standardized re-annotated version of the released dataset. Both the open-source interface and the novelly annotated dataset are intended to foster broader contributions and future works. The demonstrated stable performance across diverse plume morphologies and challenging imaging conditions, suggests strong generalization potential beyond volcanology. Applicability may extend to plume-shaped phenomena across a wide range of environmental contexts, including aeolian dust remobilization, atmospheric clouds, wildfire smoke, but also anthropogenic plumes such as smokestack emissions and industrial fires.

\subsubsection{Acknowledgments} 

This work has been supported by the Alliance Campus Rhodanien (ACR) through the 2024 call for projects (VolCloud-AI). This research has also been supported by the Swiss National Science Foundation (grant no. 200020\_204315/1).

\bibliographystyle{splncs04_2authors_etal}
\bibliography{references}

\begin{thebibliography}{10}
\providecommand{\url}[1]{\texttt{#1}}
\providecommand{\urlprefix}{URL }
\providecommand{\doi}[1]{https://doi.org/#1}

\bibitem{Aravena2025}
Aravena, A., et~al.: {{PhixCam}}: {{A Tool}} to {{Georeference Images Captured}} by {{Visible Cameras}} with {{Applications}} for {{Volcano Monitoring}}. Remote Sensing  \textbf{17}(21) (2025)

\bibitem{Benenson_2019_CVPR}
Benenson, R., et~al.: Large-scale interactive object segmentation with human annotators. In: CVPR. pp. 11700--11709 (2019)

\bibitem{berg2019ilastik}
Berg, S., et~al.: Ilastik: interactive machine learning for (bio) image analysis. Nature methods  \textbf{16}(12),  1226--1232 (2019)

\bibitem{Bommasani2022}
Bommasani, R., et~al.: On the opportunities and risks of foundation models. arXiv preprint arXiv:2108.07258  (2021)

\bibitem{carion2026sam}
Carion, N., et~al.: {SAM} 3: Segment anything with concepts. In: ICLR (2026)

\bibitem{centeno2024near}
Centeno, R., et~al.: Near-real-time multiparametric seismic and visual monitoring of explosive activity at sabancaya volcano, peru. Journal of Volcanology and Geothermal Research  (2024)

\bibitem{Chen2017}
Chen, Q., et~al.: Fast image processing with fully-convolutional networks. In: ICCV. pp. 2497--2506 (2017)

\bibitem{cheng2021mask2former}
Cheng, B., et~al.: Masked-attention mask transformer for universal image segmentation. In: CVPR. pp. 1290--1299 (2022)

\bibitem{Dosovitskiy2021}
Dosovitskiy, A., et~al.: An image is worth 16x16 words: Transformers for image recognition at scale. In: ICLR (2021)

\bibitem{sam2_studio}
Face, H.: sam2-studio (2024), \url{https://github.com/huggingface/sam2-studio}, last accessed 2026-04-10

\bibitem{giffard2025vigia}
Giffard-Roisin, S., et~al.: Vigia-plumenet and vigia-plumedata: Open-source ai segmenting model and database for volcanic plumes and emissions from optical cameras. Volcanica  (2025)

\bibitem{guerrero2022convolutional}
Guerrero~Tello, J.F., et~al.: Convolutional neural network algorithms for semantic segmentation of volcanic ash plumes using visible camera imagery. Remote Sensing  (2022)

\bibitem{segment_anything_2_ui}
Hesko, B.: segment-anything-2-ui (2024), \url{https://github.com/branislavhesko/segment-anything-2-ui}, last accessed 2026-04-10

\bibitem{hollandi2020annotatorj}
Hollandi, R., et~al.: Annotatorj: an imagej plugin to ease hand annotation of cellular compartments. Molecular biology of the cell  \textbf{31}(20),  2179--2186 (2020)

\bibitem{Huang_2023_ICCV}
Huang, Y., et~al.: Interformer: Real-time interactive image segmentation. In: ICCV. pp. 22301--22311 (2023)

\bibitem{sam_hq}
Ke, L., et~al.: Segment anything in high quality. In: NeurIPS (2023)

\bibitem{Kirillov2023}
Kirillov, A., et~al.: Segment anything. In: ICCV. pp. 4015--4026 (2023)

\bibitem{lin2017feature}
Lin, T.Y., et~al.: Feature pyramid networks for object detection. In: CVPR. pp. 2117--2125 (2017)

\bibitem{liu2015optimising}
Liu, E., et~al.: Optimising shape analysis to quantify volcanic ash morphology. GeoResJ  \textbf{8},  14--30 (2015)

\bibitem{Long2015}
Long, J., et~al.: Fully convolutional networks for semantic segmentation. In: CVPR. pp. 3431--3440 (2015)

\bibitem{SAM2_GUI}
Mao, Y.: Sam2-gui (2024), \url{https://github.com/YunxuanMao/SAM2-GUI}, last accessed 2026-04-10

\bibitem{oquab2023dinov2}
Oquab, M., et~al.: Dinov2: Learning robust visual features without supervision. arXiv preprint arXiv:2304.07193  (2023)

\bibitem{pachitariu2025cellpose}
Pachitariu, M., et~al.: Cellpose-sam: superhuman generalization for cellular segmentation. BioRxiv pp. 2025--04 (2025)

\bibitem{Perazzi2016}
Perazzi, F., et~al.: A benchmark dataset and evaluation methodology for video object segmentation. In: Computer Vision and Pattern Recognition (2016)

\bibitem{Radford2021}
Radford, A., et~al.: Learning transferable visual models from natural language supervision. In: ICML. pp. 8748--8763. PmLR (2021)

\bibitem{Ravi2024}
Ravi, N., et~al.: {SAM} 2: Segment anything in images and videos. In: ICLR (2025)

\bibitem{Ronneberger2015}
Ronneberger, O., et~al.: U-net: Convolutional networks for biomedical image segmentation. In: MICCAI. pp. 234--241. Springer (2015)

\bibitem{schneider2012nih}
Schneider, C.A., et~al.: Nih image to imagej: 25 years of image analysis. Nature methods  \textbf{9}(7),  671--675 (2012)

\bibitem{Simionato2024a}
Simionato, R., et~al.: Combining {{Visible-}} and {{Infrared-Wavelength Observations With Numerical Modeling}} to {{Describe Vulcanian Eruption Plumes}} at {{Sabancaya}}, {{Peru}}. Journal of Geophysical Research: Solid Earth  \textbf{129}(12) (2024)

\bibitem{Simionato2022a}
Simionato, R., et~al.: {{PlumeTraP}}: {{A New MATLAB-Based Algorithm}} to {{Detect}} and {{Parametrize Volcanic Plumes}} from {{Visible-Wavelength Images}}. Remote Sensing  \textbf{14}(7) (2022)

\bibitem{sofiiuk2022reviving}
Sofiiuk, K., et~al.: Reviving iterative training with mask guidance for interactive segmentation. In: ICIP. pp. 3141--3145. IEEE (2022)

\bibitem{Sultana2020}
Sultana, F., et~al.: Evolution of {{Image Segmentation}} using {{Deep Convolutional Neural Network}}: {{A Survey}}. Knowledge-Based Systems  \textbf{201--202},  106062 (2020)

\bibitem{Szeliski2022}
Szeliski, R.: Computer {{Vision}}: {{Algorithms}} and {{Applications}}. Texts in {{Computer Science}}, Springer International Publishing, Cham (2022)

\bibitem{valade2014plume}
Valade, S., et~al.: Plume ascent tracker: Interactive matlab software for analysis of ascending plumes in image data. Computers \& Geosciences  \textbf{66},  132--144 (2014)

\bibitem{Wilkes2022}
Wilkes, T.C., et~al.: Semantic segmentation of explosive volcanic plumes through deep learning. Computers \& Geosciences  \textbf{168},  105216 (2022)

\bibitem{wong2024scribbleprompt}
Wong, H.E., et~al.: Scribbleprompt: fast and flexible interactive segmentation for any biomedical image. In: ECCV. pp. 207--229. Springer (2024)

\bibitem{xie2024moving}
Xie, J., et~al.: Moving object segmentation: All you need is sam (and flow). In: ACCV. pp. 162--178 (2024)

\bibitem{zou2023segment}
Zou, X., et~al.: Segment everything everywhere all at once. NeurIPS  \textbf{36},  19769--19782 (2023)

\end{thebibliography}

\newpage

\end{document}